\documentclass[Afour,sageh,times]{sagej}

\usepackage{moreverb,url}
\usepackage{amsmath} % assumes amsmath package installed
\usepackage{amssymb}  % assumes amsmath package installed
\usepackage{subfigure}
\usepackage{multirow}
\usepackage{array,booktabs}
\usepackage{diagbox}
\usepackage{balance}
\usepackage{mathtools}
\usepackage{subfigure}
\usepackage{verbatim}
\usepackage{adjustbox}
\usepackage{algorithm}
\usepackage{algpseudocode}
\usepackage{romannum}
\usepackage{amssymb}% http://ctan.org/pkg/amssymb
\usepackage{pifont}% http://ctan.org/pkg/pifont
\usepackage{hyperref}
\usepackage{enumitem}
\hypersetup{
  colorlinks=true, %Colours links instead of ugly boxes
  urlcolor=blue,   %Colour for external hyperlinks
  linkcolor=blue,  %Colour of internal links
  citecolor=black  %Colour of citations
}
\usepackage[table]{xcolor}
\usepackage{./rpm_packages/rpm_SIunits}
\usepackage{./rpm_packages/rpm_acronyms}
\usepackage{./rpm_packages/rpm_math}
\usepackage{./rpm_packages/rpm_misc}
\usepackage{soul,color}
\usepackage{multirow}
\usepackage{soul,color}
\usepackage{lipsum}
\usepackage{tikz} % \checkmark
\usepackage{MnSymbol}

\newcommand{\bl}[1]{{\textcolor{black}{#1}}}

\newcommand\BibTeX{{\rmfamily B\kern-.05em \textsc{i\kern-.025em b}\kern-.08em
T\kern-.1667em\lower.7ex\hbox{E}\kern-.125emX}}

\def\volumeyear{2024}

\usepackage{bbding,pifont}

\usepackage{xspace}
\newcommand{\eg}{e.g.\xspace}

\begin{document}

\runninghead{Jang \emph{et al.}}
% \runninghead{TRansPose: Large-Scale Multispectral Transparent Object Dataset}

\title{MOANA: Multi-Radar Dataset for Maritime Odometry and Autonomous Navigation Application}

\author{Hyesu Jang\affilnum{1}\affilnum{\dagger}, Wooseong Yang\affilnum{2}\affilnum{\dagger}, Hanguen Kim\affilnum{3}, Dongje Lee\affilnum{3}, Yongjin Kim\affilnum{3}, Jinbum Park\affilnum{3}, Minsoo Jeon\affilnum{3}, Jaeseong Koh\affilnum{3}, Yejin Kang\affilnum{3}, Minwoo Jung\affilnum{2}, Sangwoo Jung\affilnum{2}, \bl{Chng Zhen Hao\affilnum{4}, Wong Yu Hin\affilnum{4}, Chew Yihang\affilnum{4}}, and Ayoung Kim\affilnum{2}}
% \author{Alistair Smith\affilnum{1} and Hendrik Wittkopf\affilnum{2}}

\affiliation{\affilnum{1} Institute of Advanced Machines and Design, SNU, Seoul, S. Korea \\
\affilnum{2} Dept. of Mechanical Engineering, SNU, Seoul, S. Korea \\
\affilnum{3} Seadronix, Seoul, S. Korea\\
\affilnum{4} Defence Science and Technology Agency, Singapore\\
\affilnum{\dagger}Two authors contributed equally.}

\corrauth{Ayoung Kim, Dept. of Mechanical Engineering, SNU, Seoul, S. Korea}
\email{ayoungk@snu.ac.kr}

\begin{abstract}
Maritime environmental sensing requires overcoming challenges from complex conditions such as harsh weather, platform perturbations, large dynamic objects, and the requirement for long detection ranges. While cameras and LiDAR are commonly used in ground vehicle navigation, their applicability in maritime settings is limited by range constraints and hardware maintenance issues. Radar sensors, however, offer robust long-range detection capabilities and resilience to physical contamination from weather and saline conditions, making it a powerful sensor for maritime navigation.
Among various radar types, X-band radar (\eg, marine radar) is widely employed for maritime vessel navigation, providing effective long-range detection essential for situational awareness and collision avoidance. Nevertheless, it exhibits limitations during berthing operations where near-field detection is critical. To address this shortcoming, we incorporate W-band radar (\eg, Navtech imaging radar), which excels in detecting nearby objects with a higher update rate.
We present a comprehensive maritime sensor dataset featuring multi-range detection capabilities. This dataset integrates short-range LiDAR data, medium-range W-band radar data, and long-range X-band radar data into a unified framework. Additionally, it includes object labels for oceanic object detection usage, derived from radar and stereo camera images. The dataset comprises seven sequences collected from diverse regions with varying levels of \bl{navigation algorithm} estimation difficulty, ranging from easy to challenging, and includes common locations suitable for global localization tasks.
This dataset serves as a valuable resource for advancing research in place recognition, odometry estimation, SLAM, object detection, and dynamic object elimination within maritime environments.
Dataset can be found in following link: \texttt{\url{https://sites.google.com/view/rpmmoana}}

\end{abstract}

% marine radar에서 엄밀히 따지면 S-band 와 X-band레이더가 사용되고있는데, pohang canal dataset에서는 maritime radar를 사용한다고 한다음에 specification에 X-band라고만 써놓았습니다.
% W-band radar가 navtech에서 X-band레이더의 단점을 커버하기 위해 사용한것이라고 나와있어서 navtech radar라고는 써도 될 것 같습니다.
%By fusing data from multiple sensors—including LiDAR for short-range detection, W-band radar for medium-range detection, and X-band radar for long-range detection—we establish the high-precision autonomous sailing systems.

\keywords{Dataset, Maritime, Radar, LiDAR, Object Detection, Place Recognition, Odometry, SLAM}

\maketitle

\section{Introduction}
As autonomous vehicles gain prominence within the field of robotics, the demand for research and development continues to rise. Many advancements in this domain have been driven by high-quality datasets featuring well-calibrated sensor configurations and carefully designed trajectories. From ground-driving datasets \citep{Geiger2012CVPR} to aerial datasets, and from imaging sensors to range sensors \citep{kim2020mulran}, various datasets have been made publicly available to support robust advancements in autonomous systems.

However, exploring oceanic environments remains a significant challenge. While high-quality datasets are essential for impactful research, the availability of maritime datasets lags behind current demands due to the complexities associated with sensor configuration and the inherent challenges of data collection in maritime environments.

The existing maritime dataset \citep{chung2023pohang} represents a significant contribution, providing navigational data from marine radar only with X-band wavelength, along with \ac{LiDAR}, camera systems, and ground truth information. While this dataset marks a pioneering step for maritime research, it lacks regional diversity, limiting its applicability across varied navigational environments.
% Moreover, the dataset faces limitations in sensor capabilities, as the canal environment is insufficient to represent radar-based marine navigation.
Understanding this limitation requires familiarity with the stages of vessel navigation: berthing at a port, sailing in open waters, and docking. While near-range detection is critical during berthing and docking, long-range detection is essential for open-water navigation. Existing datasets provide limited coverage of such wide-open maritime areas, restricting utility for radar navigation research.
\begin{figure*}[!t]
    \centering
    \includegraphics[width=0.89\linewidth]{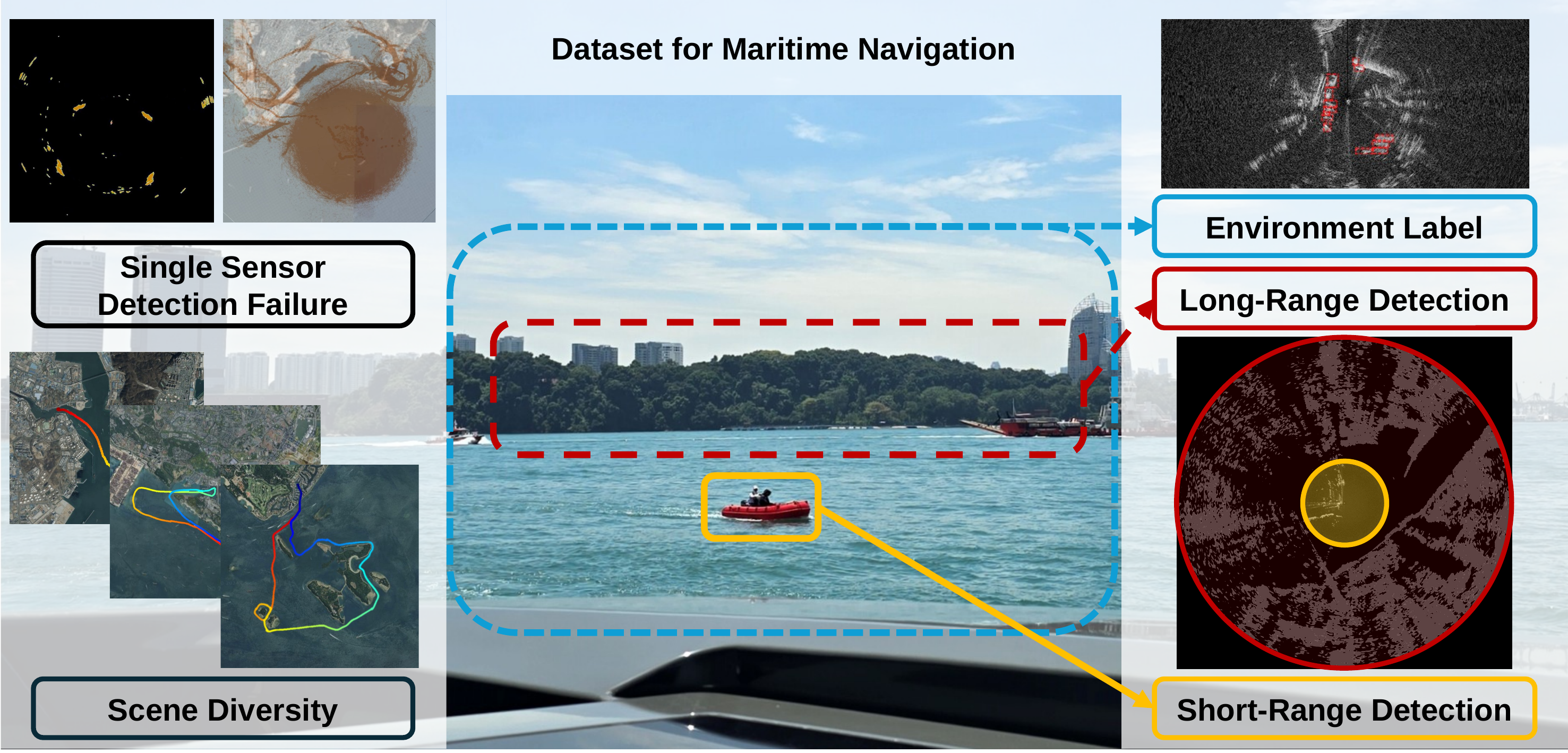}
    \vspace{-2mm}
    \caption{Overview of the MOANA dataset. To address the limitations of individual sensor performance in maritime environments, we enhanced the sensor system by integrating complementary sensors to improve both range and resolution. Data acquisition was carried out across diverse scenarios using multiple sensor types, providing a robust dataset with annotated labels for the development of learning-based algorithms.}
    \label{fig:overview}
    \vspace{-5mm}
\end{figure*}

Moreover, the existing dataset faces limitations in sensor capabilities, \bl{as the canal environment is the narrow area generating multipath effects in radar imagery, which is insufficient to represent radar-based navigation}. Traditional marine radars (X-band) are pivotal in ensuring situational awareness and preventing collisions in wide oceanic areas with their long-range detection ability.
Despite its broad sensing range, X-band radars are limited in detecting \bl{small objects near the vessels} and often suffer from multipath interference and noise, as illustrated in \figref{fig:overview}.
Therefore, it is necessary to explore alternative sensors for \bl{these types of} detection.
While LiDAR has proven \bl{its effectiveness for the high-resolution perception}, it is not a sustainable solution in maritime environments due to challenges like sea fog, corrosion, and the necessity for frequent cleaning during extended voyages.
%우리는 short ragne 와 wide area detection에서 lidar를 대체하고 marine 환경에 W-band radar 가 굉장히 적합함을 발견하고 최초로 marine 환경에 w-band 센서 데이터를 포함한 데이터를 출간한다.
% W-band는 기존에는 모두 ground vehicle 에만 적용되어 그 가능성을 보였으나 해상환경에서 데이터가 보고된바가 없다
We have found that W-band radar can be a superior alternative to LiDAR for \bl{short-range} detection in marine environments.
W-band radar, often called scanning radar, has previously demonstrated its potential exclusively in ground vehicle applications \citep{harlow2024new} due to its extended detection range \bl{and robustness to the surroundings} compared to LiDAR. \bl{Its environmentally resilient detection capabilities can effectively address the shortcomings of LiDAR in the maritime domain. Also, its higher resolution relative to the X-band radar provides comprehensive perception ability, which can be a reasonable alternative for LiDAR in short-range detection with high fidelity.} Nevertheless, there have been no reports of its use in marine environments.
% However, in the maritime domain, W-band radar presents a comparatively short detection range. Despite its high resolution, the W-band radar becomes less effective in open waters, where objects and structures are typically located at greater distances.

Overcoming the aforementioned limitations in the existing marine dataset, we present a maritime navigational dataset with a novel sensor configuration tailored for various maritime environments. We integrated multiple radar systems into the dataset, combining W-band radar with X-band radar to ensure comprehensive coverage. This novel configuration enables us to address both long-range and short-range detection challenges, enhancing the data continuity for diverse maritime navigation scenarios.
The expected influences of our dataset are written below:

\begin{enumerate}
    \item The MOANA dataset represents the first multi-radar (X-band and W-band) dataset collected in a maritime environment. The X-band radar is utilized for wide-area detection, while the W-band radar delivers high-resolution imaging for near-port areas.  Their differing bandwidths allow for seamless integration, and our benchmark results demonstrate that a complementary sensor configuration enables robust navigation across various maritime conditions.
    Our dataset also integrates cameras and LiDAR data, encouraging diverse sensor fusion strategies for vessel navigation tasks: berthing, sailing, and docking. \bl{Moreover, the ground truth labels of detected objects for each modality are also provided for facilitating rigorous validation and benchmarking of multi-modal object detection in oceanic areas.} 
    
    \item The MOANA dataset tackles maritime navigation tasks in varying environments. Our dataset encompasses both structured port and unstructured ocean and island settings with varying vessel sizes. This large spectrum of scene diversity underscores the necessity of navigation algorithms operating in multifarious environments. Additionally, the loops and overlaps between sequences support inter- and intra-place recognition in maritime environments.
    
    \item The MOANA dataset provides a series of challenges designed to evaluate the robustness of existing navigation algorithms. The sequences include objectless data, ghost detections caused by multipath noise, and the presence of large dynamic objects. All are aimed at preventing unpredictable conditions that can affect the robustness of navigation systems.

\end{enumerate}
\section{Related works}

% \usepackage{multirow}
% \usepackage{graphicx}
% \usepackage{multirow}
% \usepackage{graphicx}

\hypersetup{
    % colorlinks=true,
    citecolor=red
}

\begin{table*}[]
\centering
\caption{Radar Dataset Comparison}
\label{tab:dataset}
\resizebox{\textwidth}{!}{%
\begin{tabular}{c|ccc|cc|cc|cc} \toprule
\multirow{2}{*}{Name} &
  \multicolumn{3}{c|}{Radar} &
  \multicolumn{2}{c|}{Navigation} &
  \multicolumn{2}{c|}{Object Detection} &
  \multirow{2}{*}{Route Complexity}&
  \multirow{2}{*}{Environment} \\ 
 &
  W-Band &
  X-Band &
  4D &
  Loop Closure &
  Scene Diversity&
  Label &
  Ground Truth &
   &
   \\ \midrule
\begin{tabular}[c]{@{}c@{}}Oxford Radar\\ \cite{RadarRobotCarDatasetICRA2020} \end{tabular}  &   \ding{51}&  \ding{55}&  \ding{55}&  \ding{51}&  Identical& \ding{55}&  \ding{55}&  $\filledstar$ $\filledstar$ $\filledstar$& Land\\
\begin{tabular}[c]{@{}c@{}}MulRan\\ \cite{kim2020mulran} \end{tabular}                            &   \ding{51}&  \ding{55}&  \ding{55}&  \ding{51}& $\filledstar$ $\filledstar$& \ding{55}& \ding{55} & $\filledstar$ $\filledstar$ $\filledstar$ & Land \\
\begin{tabular}[c]{@{}c@{}}Radiate\\ \cite{sheeny2021radiate} \end{tabular}                             &   \ding{51}&  \ding{55}&  \ding{55}&  \ding{51}& $\filledstar$ $\filledstar$& \ding{51}& \ding{51}& $\filledstar$ $\filledstar$ &  Land \\
\begin{tabular}[c]{@{}c@{}}Boreas\\ \cite{burnett2023boreas} \end{tabular}                             &   \ding{51}&  \ding{55}&  \ding{55}&  \ding{51}& $\filledstar$& \ding{51}& \ding{51}& $\filledstar$ $\filledstar$ &  Land \\
\begin{tabular}[c]{@{}c@{}}OORD\\ \cite{gadd2024oord} \end{tabular}                             &   \ding{51}&  \ding{55}&  \ding{55}&  \ding{51}& $\filledstar$ $\filledstar$ $\filledstar$& \ding{55}& \ding{55}& $\filledstar$ $\filledstar$ &  Land \\
\begin{tabular}[c]{@{}c@{}}HeRCULES\\ \cite{kim2025hercules} \end{tabular}                             &   \ding{51}&  \ding{55}&  \ding{51}&  \ding{51}& $\filledstar$ $\filledstar$ $\filledstar$& \ding{55}& \ding{55}& $\filledstar$ $\filledstar$ &  Land \\ \midrule

\begin{tabular}[c]{@{}c@{}}USVInland\\ \cite{cheng2021we} \end{tabular}                            &   \ding{55}&  \ding{55}&  \ding{51}&  \ding{51}& $\filledstar$ $\filledstar$ $\filledstar$& \ding{51}(Water)& \ding{51} & $\filledstar$ $\filledstar$ & Waterway \\
\begin{tabular}[c]{@{}c@{}}Pohang Canal\\ \cite{chung2023pohang} \end{tabular}                       &   \ding{55}&  \ding{51}&   \ding{55} &  \ding{55}&  Identical&  \ding{55}& \ding{55}& $\filledstar$ & Maritime  \\ \midrule
\textbf{MOANA}                                                                             &  \textbf{ \ding{51}}&  \textbf{\ding{51}}&  \textbf{\ding{55}} &  \ding{51}&  $\filledstar$ $\filledstar$ $\filledstar$ & \ding{51}&   \ding{51} & $\filledstar$ $\filledstar$ $\filledstar$ & Maritime \\ \bottomrule
\end{tabular}%
}
\end{table*}

\hypersetup{
    % colorlinks=true,
    citecolor=black
}

%The advent of large-scale camera and LiDAR datasets has significantly enhanced the generalization capabilities of autonomous vehicle systems. With advancements in computing technology, the demand for high-quality sensor data has also risen.

% \textcolor{blue}{Compared to camera or LiDAR, radar has attracted increasing attention for autonomous driving in recent times.} Due to its longer wavelength compared to other sensors, radar offers advantages such as the ability to penetrate obstacles like buildings and operate over longer ranges. \bl{This section briefly summarizes two main imaging radar types widely used in robotics, namely X-band radars and W-band radars.}
% This section summarizes the existing maritime dataset
% focused on cameras and LiDAR and examines W-band datasets that have so far been limited to ground-based usage in advance of introducing the first multi-radar dataset to incorporate the W-band radar from maritime environments. The brief summarization is depicted in \tabref{tab:dataset}.
\bl{Prior to introducing our first multi-radar dataset MOANA, we summarize existing maritime datasets—primarily focused on cameras and LiDAR—as well as W-band radar datasets, which have been limited to ground vehicle applications. A summary is provided in Table 1.}

\subsection{Marine Robotics Datasets}

The development of maritime environment datasets also began with vision-based approaches for object detection and segmentation tasks. Early efforts included a camera-infrared dataset for day and low-light conditions \citep{zhang2015vais} and a bird's-eye-view maritime surveillance camera dataset \citep{ribeiro2017data}. Additionally, MaSTr1325 \citep{bovcon2019mastr1325} provided maritime imagery for learning-based models.
\bl{Extending beyond the image domain, multimodal datasets for sensor fusion have been presented.
FloW \citep{cheng2021flow} utilized a single-chip radar and cameras for object detection in the waterway. \citep{lin2022maritime} exploited LiDAR and \ac{AIS} for detection and semantic mapping in the maritime environment.}

Recently, datasets for maritime navigation have been presented following the increasing demand in this field.
\bl{USVInland} \citep{cheng2021we} attempted to integrate camera, LiDAR, and single-chip radar on unmanned surface vehicles, but the single-chip radar system has insufficient resolution for navigation in oceanic environments.
\bl{RoboWhaler \citep{defilippo2021robowhaler} equipped \ac{INS}, LiDAR with marine radar in \ac{ASV} for navigation applications. MassMIND \citep{nirgudkar2023massmind} integrated the long wave infrared camera on the RoboWhaler system to complement the lack of diversity of optical images in maritime environments.}
\bl{The Pohang Canal dataset} \citep{chung2023pohang} introduced a multi-purpose marine radar dataset. While camera and LiDAR data exhibit limitations in open water scenarios due to detection range constraints, including X-band marine radar in this dataset addresses these shortcomings, offering a more comprehensive solution for maritime applications.
\bl{PoLaRIS \citep{choi2024polaris} extended the Pohang Canal dataset, providing the object bounding box and tracking annotations.}
Unlike existing maritime datasets, the MOANA dataset incorporates both X-band and W-band radar to enhance the robustness of detectability for diverse maritime navigation tasks.

\subsection{W-band Radar Datasets}

% The use of W-band radar has been confined to ground vehicle datasets. Pioneering radar datasets, including MulRan \citep{kim2020mulran} and the Oxford Radar Robotcar \citep{RadarRobotCarDatasetICRA2020}, have established radar as a viable sensor for navigation.
% Following these foundational efforts, datasets have been extended to include various challenging environments, such as adverse weather conditions \citep{sheeny2021radiate}, multi-season driving \citep{burnett2023boreas}, and complex terrains \citep{gadd2024oord}.
% The MOANA dataset is differentiated from the existing W-band radar datasets, exploiting W-band radar in the maritime domain for the first time. Consequently, the MOANA dataset serves as a milestone to verify the potential of the W-band radar in vessel navigation research.

% Recently, there has been growing interest in 4D radar, though its application remains limited to short-range sensing, primarily suited for ground vehicle detection.

The use of W-band radar has been confined to ground vehicle datasets. Pioneering W-band radar datasets, including \bl{the Oxford Radar RobotCar \citep{RadarRobotCarDatasetICRA2020} and MulRan \citep{kim2020mulran} datasets, have established radar as a viable sensor for navigation. The Oxford Radar RobotCar dataset provides extensive radar data alongside complementary sensor modalities for robust localization and mapping. However, its focus on a single urban area limits environmental diversity. Similarly, the MulRan dataset offers valuable multi-modal data for urban place recognition but remains constrained by its emphasis on structured urban environments and a lack of adverse weather scenarios.}

\bl{Following these foundational efforts, Radiate \citep{sheeny2021radiate} incorporates adverse weather conditions, such as rain and snow, along with ground truth object labels, thereby demonstrating the weather-resilient perception capability of W-band radar in harsh conditions. The Boreas dataset \citep{burnett2023boreas} further expands on this by encompassing multi-seasonal data with annotation labels for object detection, showcasing radar performance under diverse seasonal conditions. Together, the Radiate and Boreas datasets highlight the potential of W-band radar for robust perception in autonomous driving applications.}

\bl{However, despite these advancements, the majority of existing W-band radar datasets remain focused on terrestrial, structured environments. Addressing this gap, OORD \citep{gadd2024oord} extends W-band radar usage from structured settings to off-road terrains, demonstrating its applicability in unstructured open environments. Recently, HeRCULES \citep{kim2025hercules} exploits W-band radar with 4D automotive radar to encompass diverse weather conditions in congested environments with high dynamics.}

\bl{The MOANA dataset is differentiated from these efforts by exploiting W-band radar in the maritime domain for the first time, thereby serving as a milestone for robust navigation in marine environments.}
\section{System Overview}

\begin{figure*}[!t]
    \centering
    \subfigure[Port sequence setup\label{fig:ulsan_setup}]{
		\includegraphics[width=0.47\linewidth]{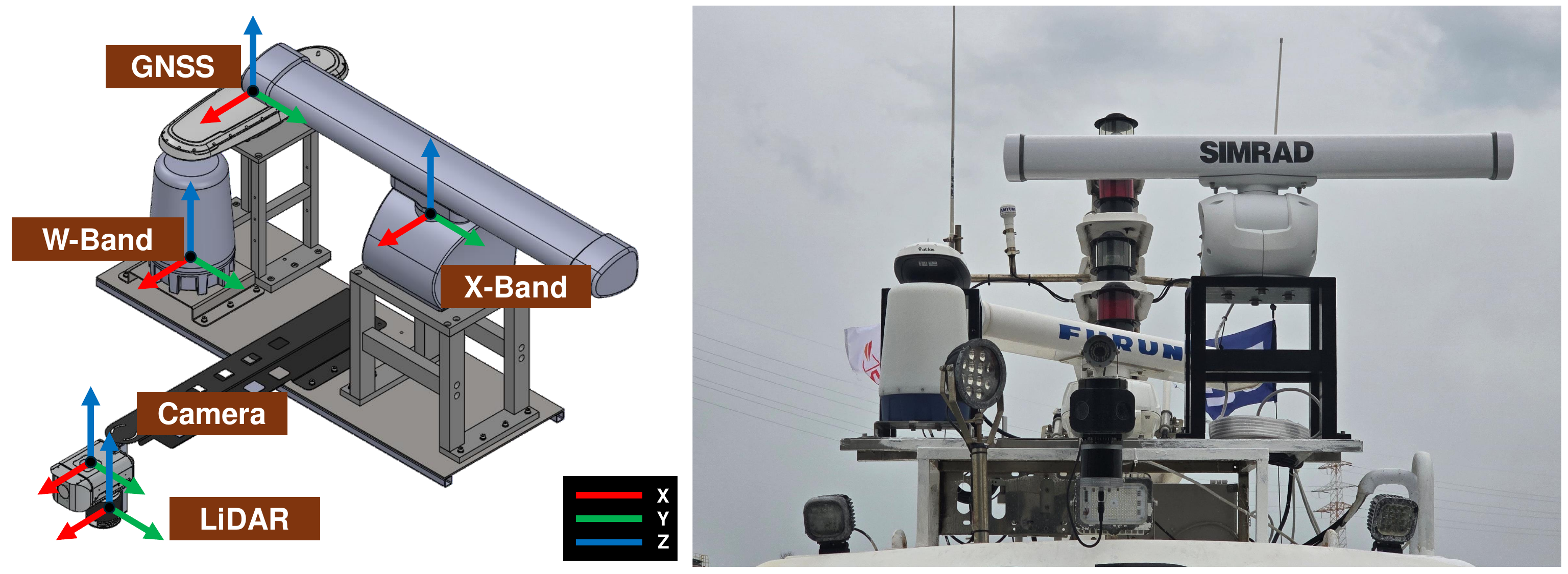}
	}
    \subfigure[Island sequence setup\label{fig:singapore_setup}]{
		\includegraphics[width=0.47\linewidth]{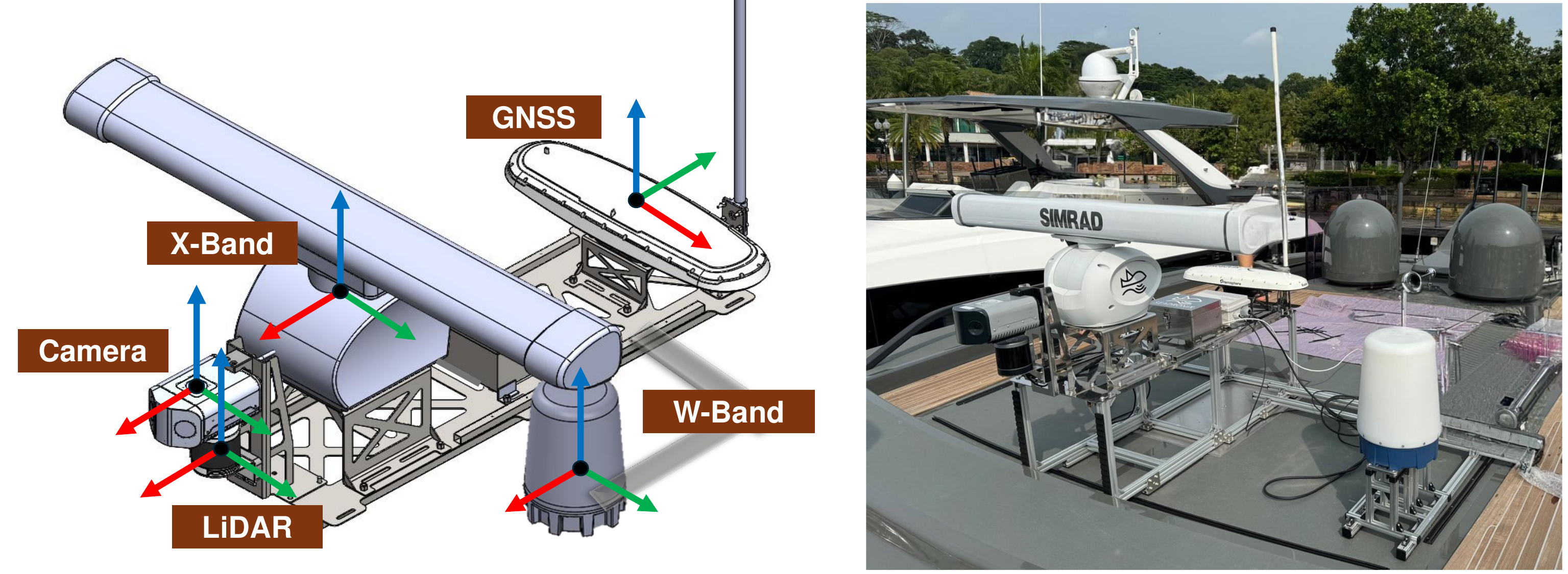}
	}
    \caption{Modeling and real-world capture setups for two distinct configurations. The primary difference between the hardware setups is the orientation of the GNSS receiver: in the port sequence, the vessel’s forward direction aligns with the x-axis, whereas the island sequence is rotated by 90°. Additionally, the W-band radar is positioned on the right side in the port sequence and on the left side in the island sequence. Detailed configurations are provided in the calibration files.}
    \label{fig:setup}
\end{figure*}

%FIGURE
\subsection{System Configuration}
\label{sec:systemConfig}
Due to the variability in conditions across different oceanic environments, identical sensor configurations could not be applied across all locations. As a result, our dataset was collected using two distinct vessels. The detailed sensor configuration is provided in \figref{fig:setup}. The first sequence, \texttt{Port}, was captured using a small fishing boat, while the \texttt{Island} sequence was recorded aboard a larger yacht. The primary sensors in the dataset are two radars: X-band and W-band. Global positioning of the system was provided by a GNSS receiver using Dual GPS mode. Additionally, two cameras and a LiDAR were integrated to support multi-modal navigational tasks. 

\subsection{Sensor Calibration}
\label{sec:sensorCalib}
The base for the yacht is established using the GNSS data. all the extrinsic calibration data are included in the calibration.
\subsubsection{Multiple Radar Calibration}
\bl{The X-band radar data is provided in the Cartesian coordinate image, whereas the W-band radar data is presented in the polar coordinate image.} We convert the W-band polar coordinate images to Cartesian coordinates with the same resolution as the X-band radar and perform the extrinsic calibration by considering the overlap of the prominent features at the pixel level. Since the two radars are structurally incapable of providing information in the vertical direction, vertical extrinsic calibration was conducted using a CAD model.

% \subsubsection{Camera-Radar Extrinsic Calibration}
% \gr{\lipsum[10]}

\subsubsection{LiDAR W-band Radar Camera Calibration}
For the LiDAR and W-band radar calibration, we utilized the phase correlation between LiDAR and W-band radar, similar to MulRan (\cite{kim2020mulran}). We first make the polar image from the bird-eye-view scan of LiDAR. Due to the limited detection range of LiDAR, we leveraged the LiDAR scan near the port area. Then, we exploited the phase correlation approach between the polar images of LiDAR and W-band radar to compute the extrinsic parameters of x, y, and yaw components. The other extrinsic parameter components were achieved with our sensor setup's CAD model.

The intrinsic calibration of the camera was performed using a known pattern target board to estimate the distortion coefficients and intrinsic parameters.
The extrinsic calibration between the camera and LiDAR was conducted through the matching of stationary planar objects with known geometries. The initial transformation for the extrinsic calibration was given by the CAD model.

%\begin{figure}[!t]
%    \centering
%    \subfigure[Canal\label{fig:canal_radar}]{
%		\includegraphics[width=0.45\columnwidth]{sections/figs/radar_bad.png}
%	}
%    \subfigure[Open water\label{fig:open_radar}]{
%		\includegraphics[width=0.45\columnwidth]{sections/figs/radar_good.png}
%	}
%    \caption{Using radar in canal is dangerous.}
%    \label{fig:canal}
%    \vspace{-5mm}
%\end{figure}

%FIGURE
\begin{figure}[!t]
    \centering
    \includegraphics[width=\columnwidth]{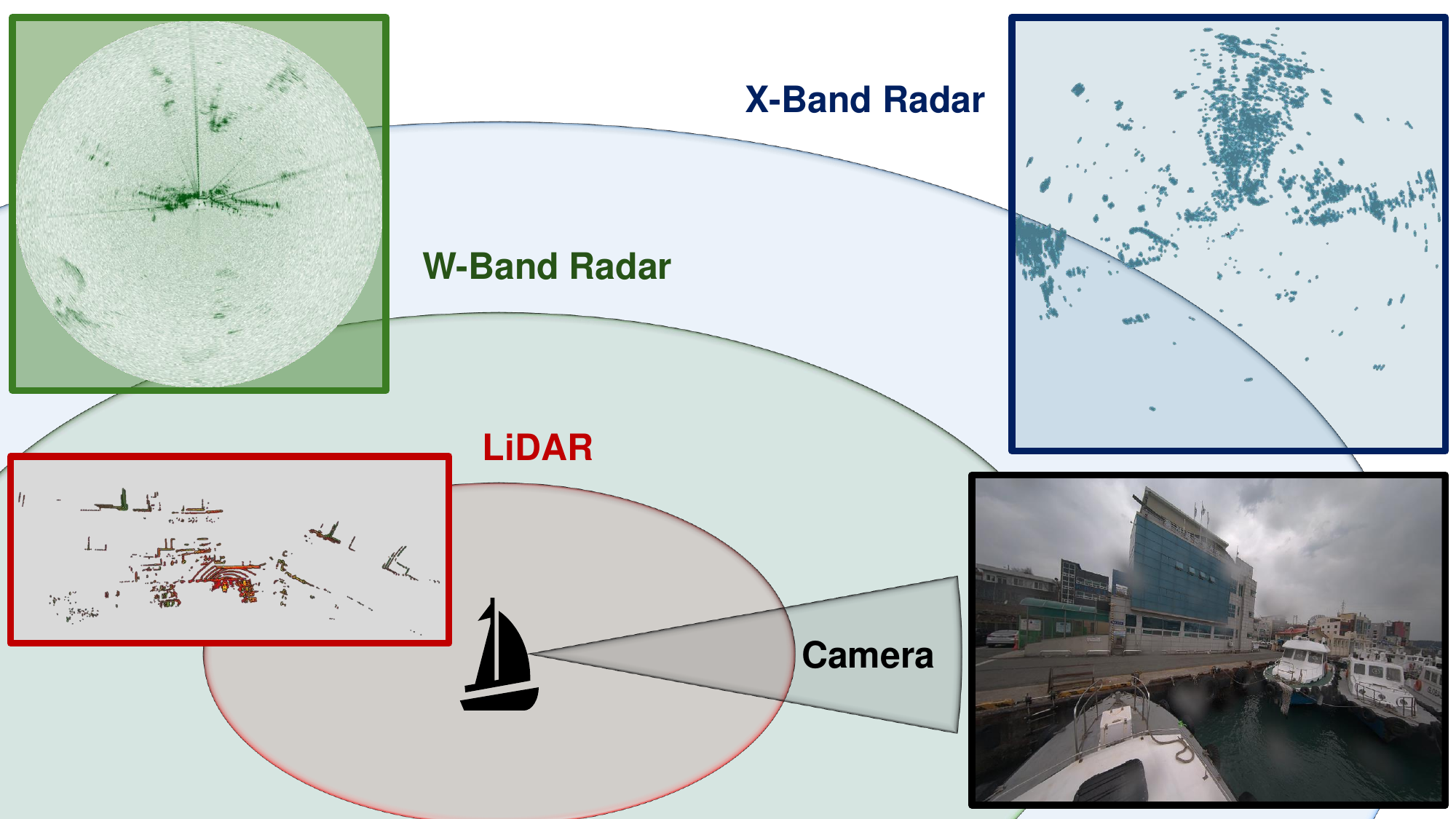}
    \caption{Composition of sensor types and example images. Radar sensors generate 360-degree scanning bird's-eye view images, while LiDAR provides 360-degree point cloud data. The camera system captures stereo images in the forward direction.}
    \label{fig:sensors}
\end{figure}

% \usepackage{multirow}
% \usepackage{graphicx}

\section{MOANA Dataset}

\subsection{Data Composition}
\begin{figure}[!t]
    \centering
    \includegraphics[width=0.95\columnwidth]{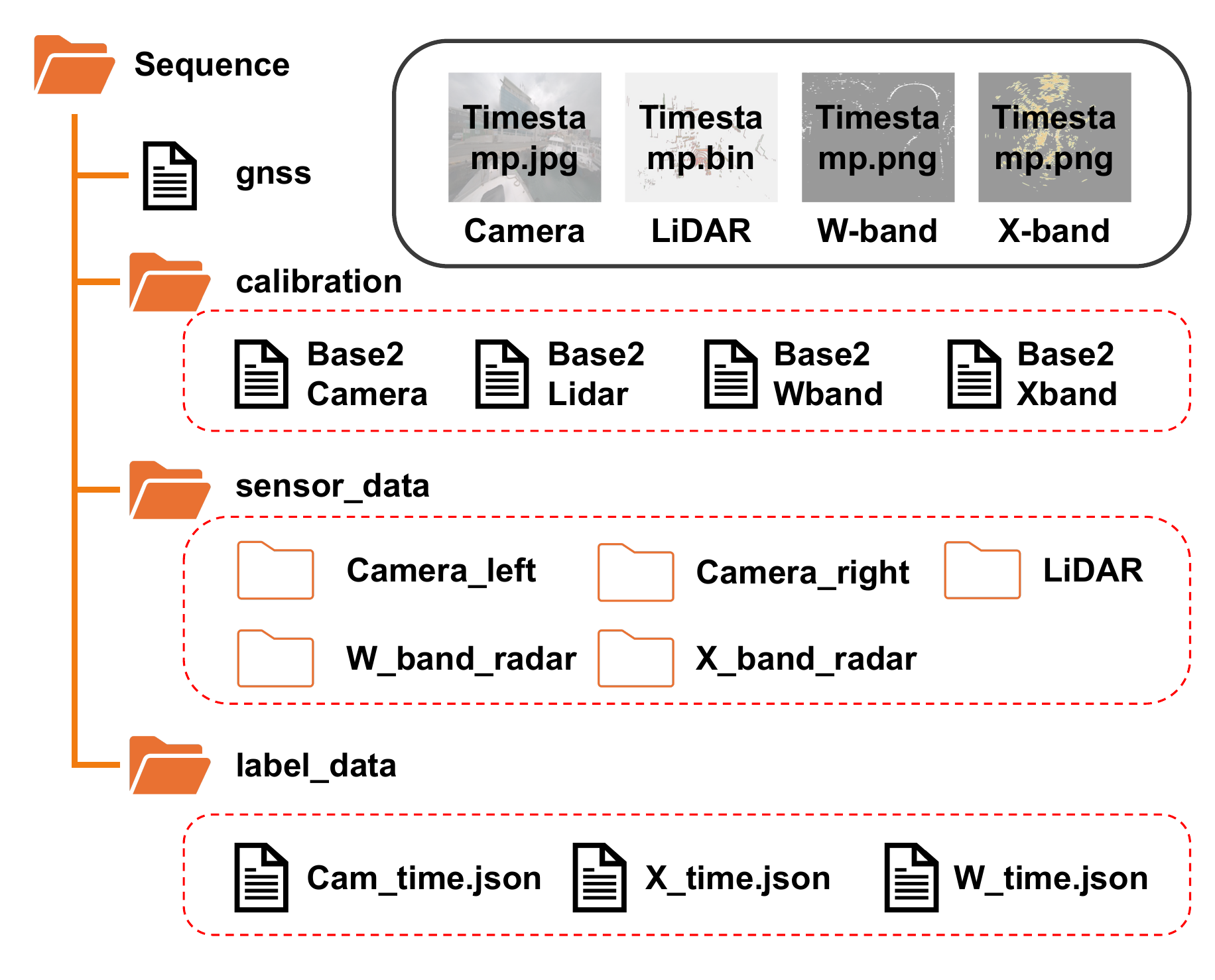}
%    \vspace{-2mm}
    \caption{Data organization and file structure for each sequence of MOANA dataset. GNSS poses are provided as text files. Extrinsic calibration parameters for the camera, LiDAR, and radars are defined relative to the base frame. Sensor data is available with distinguished file formats, such as PNG images. Additionally, labels for each frame are supplied as JSON files, each named according to the corresponding frame timestamp. For the stereo camera annotation, we denote the left camera as Cam0 and the right camera as Cam1 for JSON files.}
    \label{fig:files}
%    \vspace{-5mm}
\end{figure}

As detailed in \bl{Section} \secref{sec:systemConfig}, our dataset includes data from two different band radars, LiDAR, stereo camera, and GNSS Reciever. \figref{fig:sensors} provides the data sample with detection range comparison for each sensor. Both scanning radars and stereo cameras generate image data, making this dataset composed of four images for every data publication. To address potential challenges in managing large data volumes, we offer an individual file structure in \texttt{$Sequence/sensor\_data/sensor\_type$} directory, enabling users to download only the data relevant to their needs. Additionally, we provide a ROS-based data publisher to facilitate seamless access and integration of the dataset. An overview of the file structure is illustrated in \figref{fig:files}.

\begin{figure*}[!t]
    \centering
    \subfigure[Satellite\label{fig:sat_map}]{
		\includegraphics[height=0.23\linewidth]{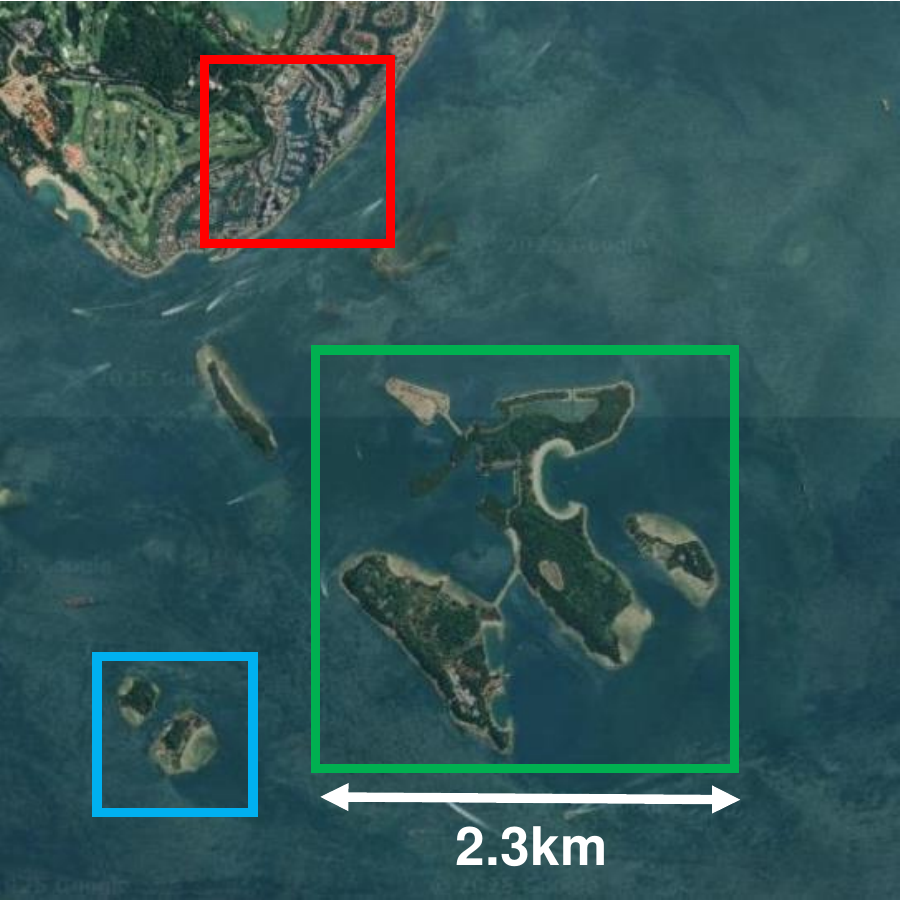}
	}
    \subfigure[X-band Radar\label{fig:x_band_map}]{
		\includegraphics[height=0.23\linewidth]{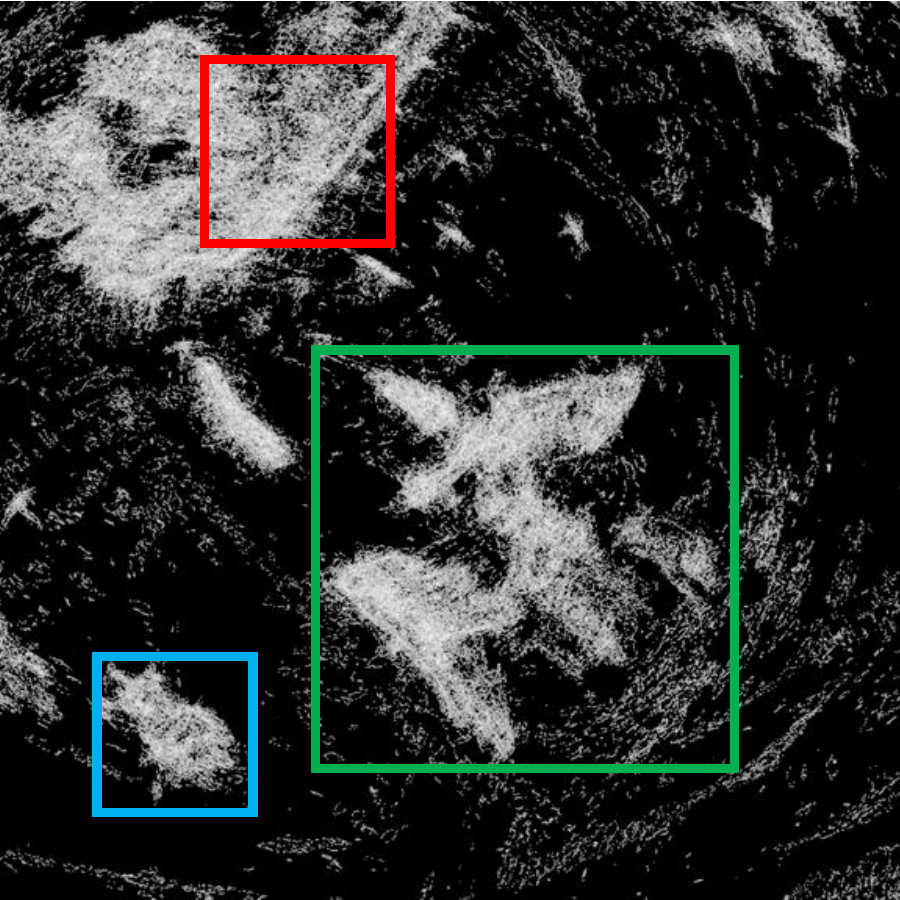}
	}
    \subfigure[W-band radar\label{fig:w_band_map}]{
		\includegraphics[height=0.23\linewidth]{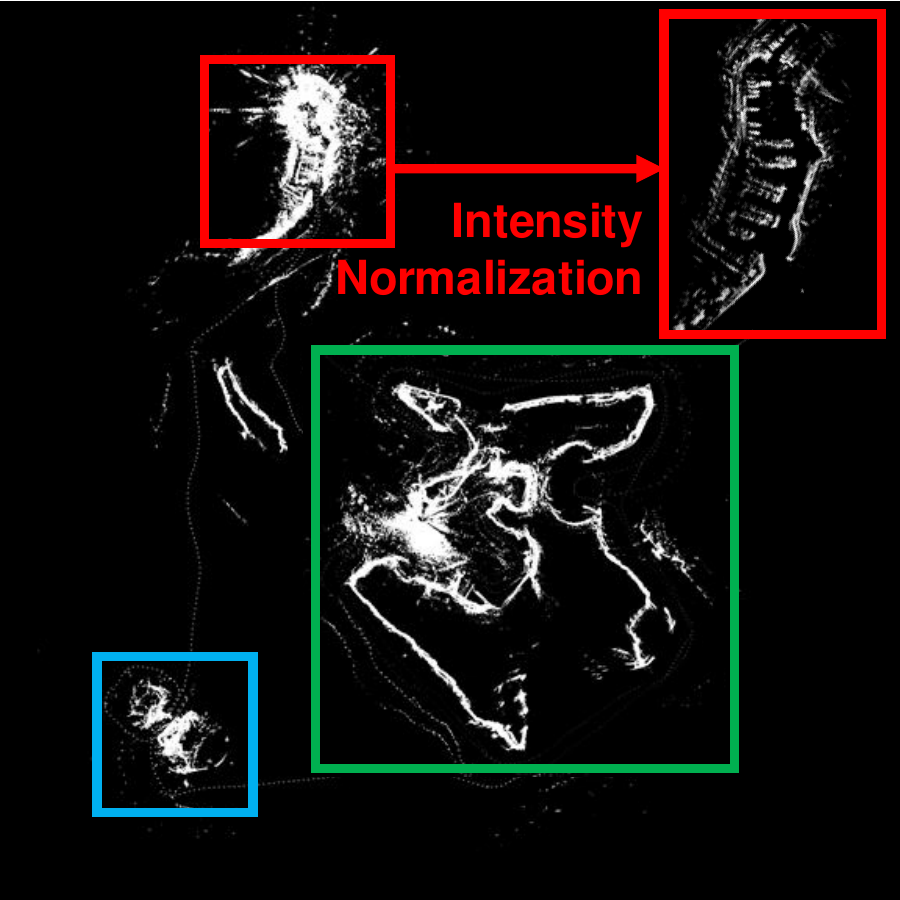}
	}
    \subfigure[LiDAR\label{fig:lidar_map}]{
		\includegraphics[height=0.23\linewidth]{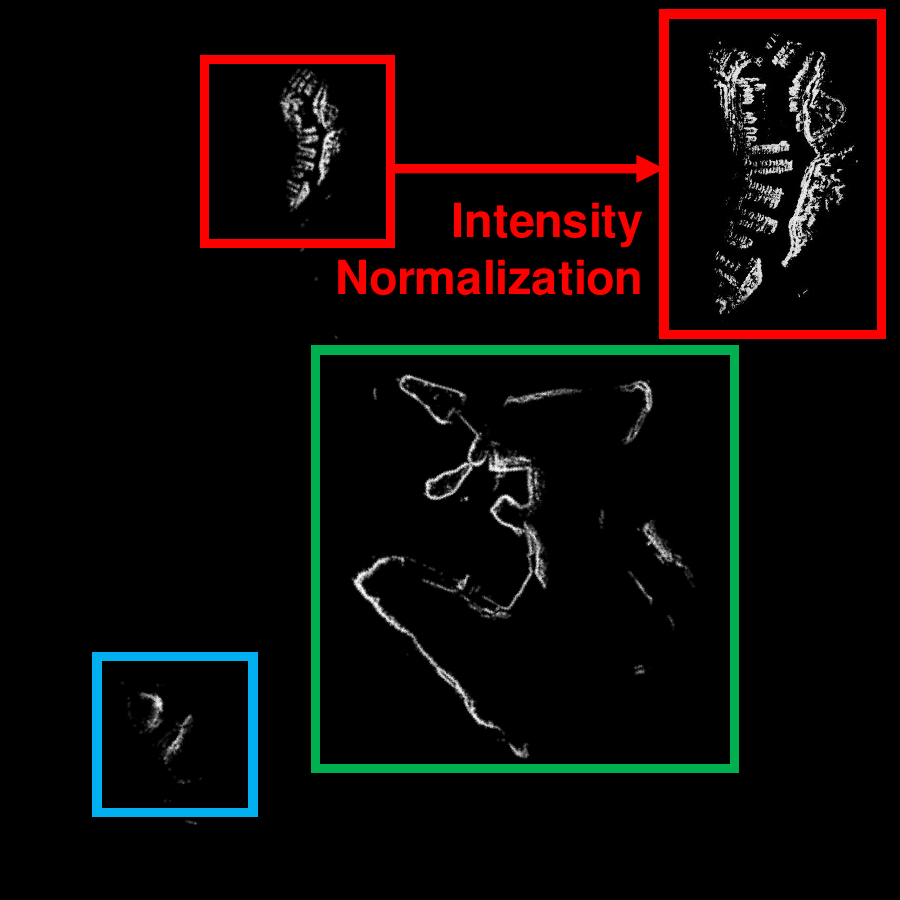}
	}
    \caption{\bl{Map constructions from GNSS data-based accumulation of the \texttt{Island} sequences. The X-band radar (b) provides full coverage of the dataset region due to its $km$ level range and strong penetration capability; however, it exhibits relatively low resolution and accuracy. In contrast, the W-band radar (c) offers high-resolution sensing but is limited to detecting only $600m$ ranges. In the berthing area (highlighted by the red-colored square), the X-band radar fails to generate a detailed map. Conversely, the W-band radar (c) and LiDAR (d) produce a high-resolution map, capturing fine structural details.}
    }
    \label{fig:maps}
\end{figure*}

\begin{figure*}[!t]
    \centering
    \subfigure[Left Camera\label{fig:annotation_cam0}]{
		\includegraphics[width=0.27\linewidth]{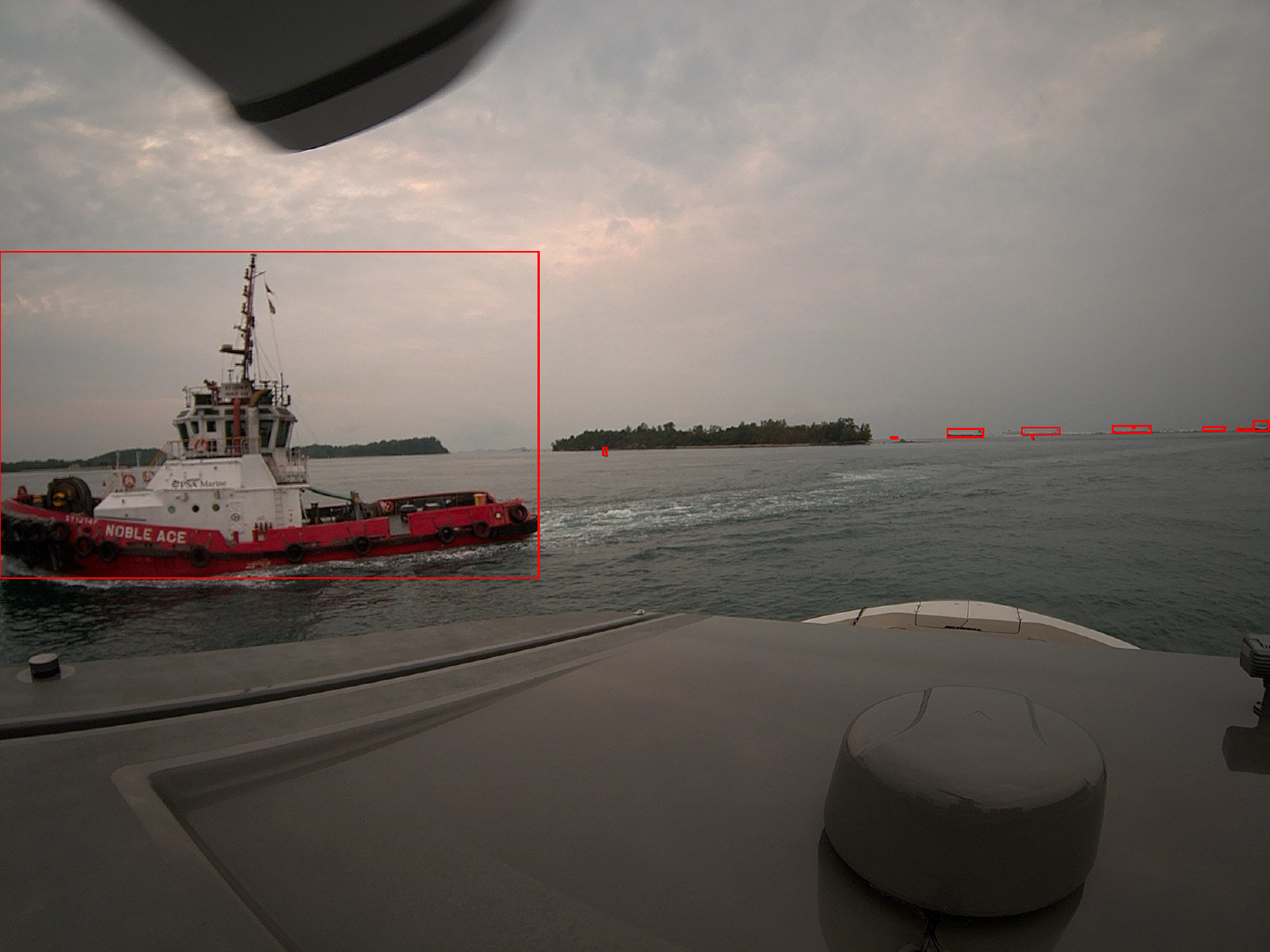}
	}
    \subfigure[Right Camera\label{fig:annotation_cam1}]{
		\includegraphics[width=0.27\linewidth]{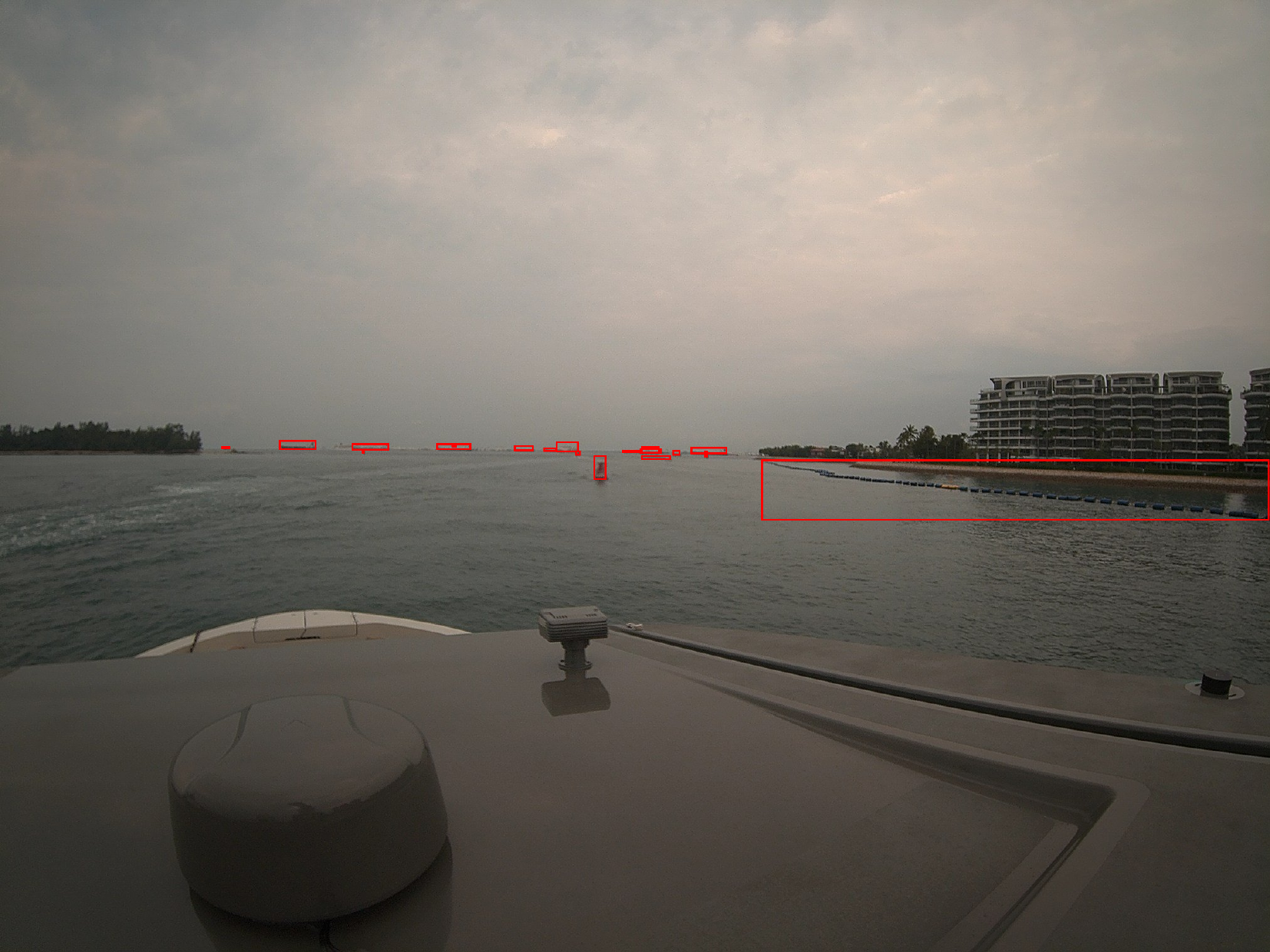}
	}
    \subfigure[X-band radar\label{fig:annotation_Xband}]{
		\includegraphics[width=0.202\linewidth]{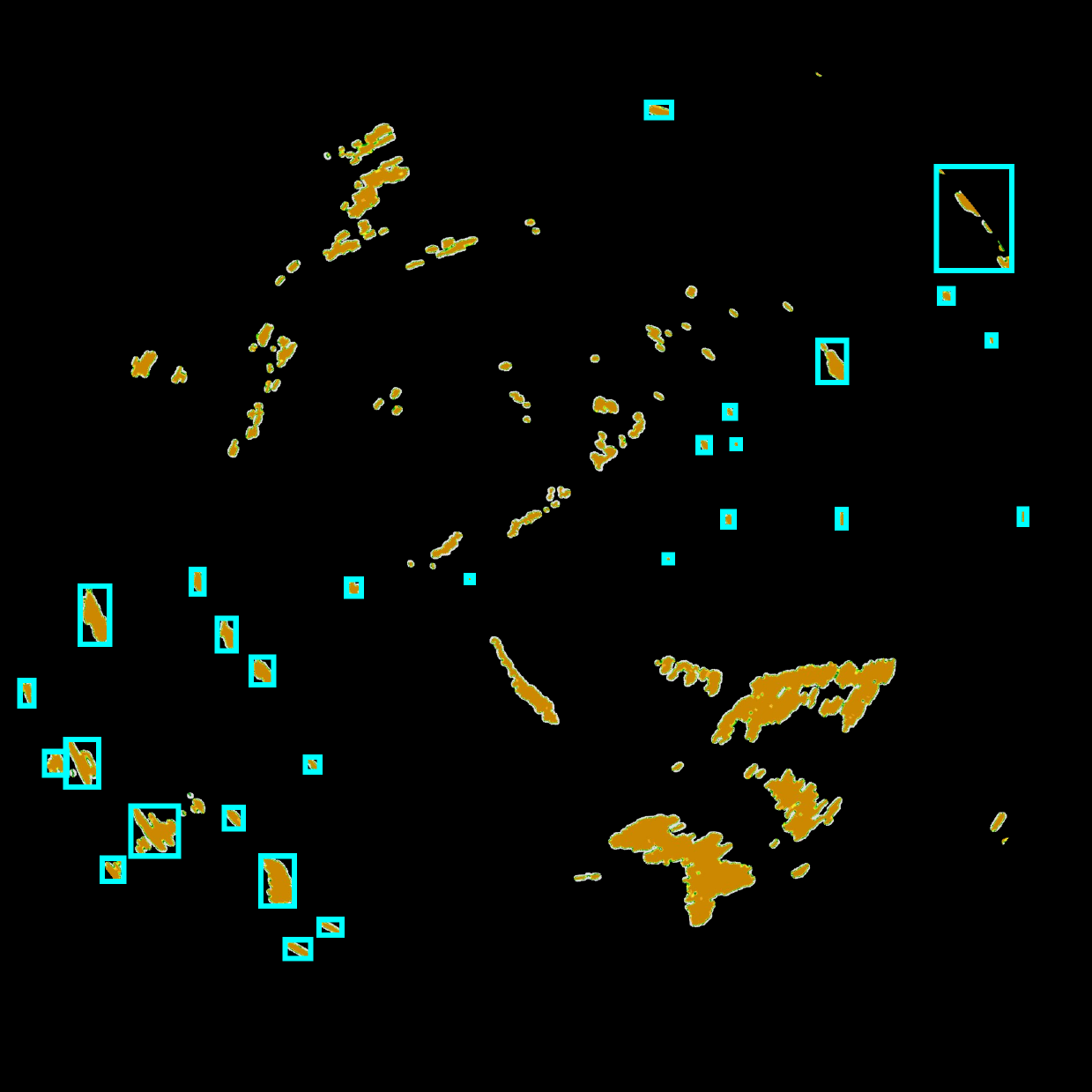}
	}
    \subfigure[W-band radar\label{fig:annotation_Wband}]{
		\includegraphics[width=0.202\linewidth]{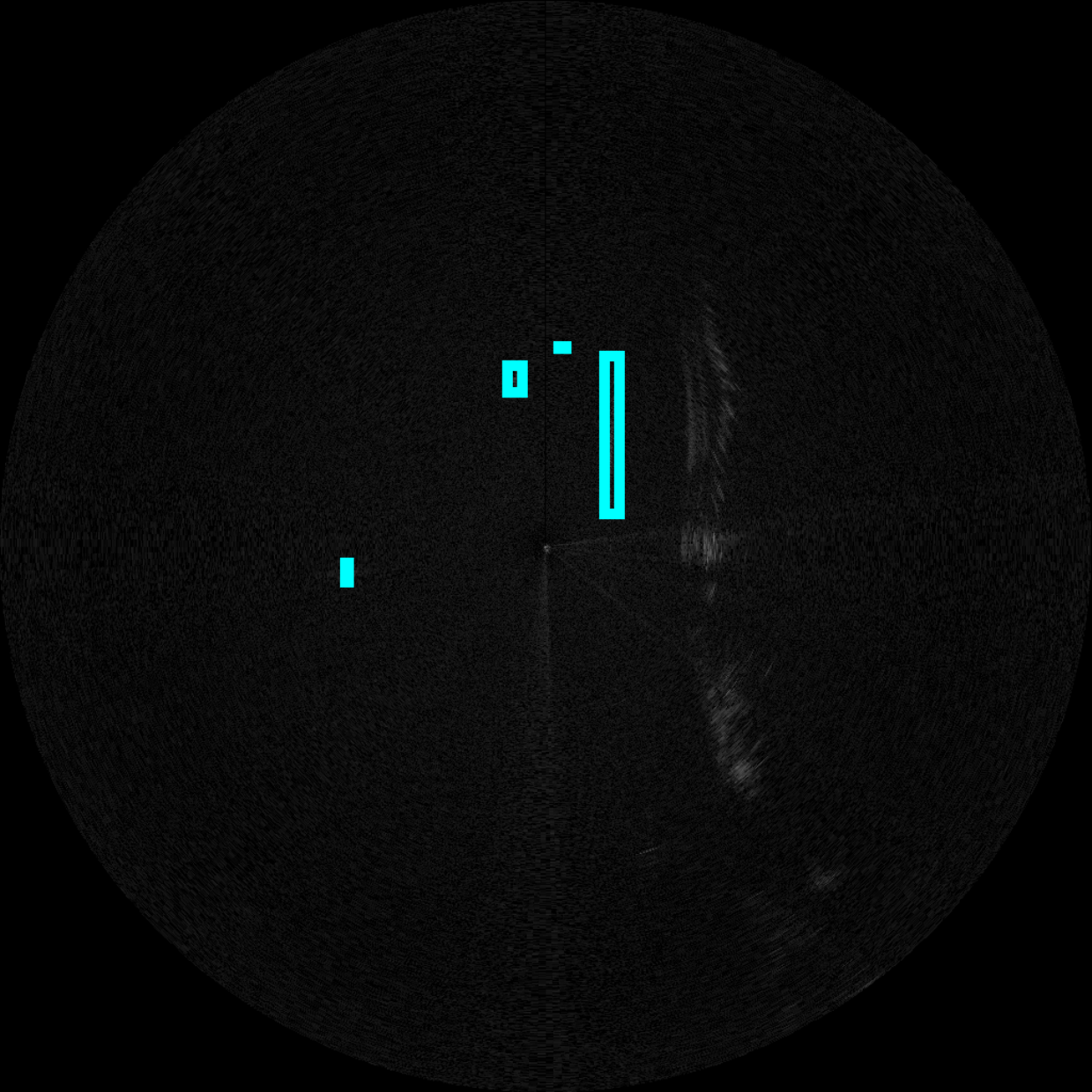}
	}
    \caption{Examples of 2D annotation for the stereo camera, X-band radar, and W-band radar data. (a,b) The bounding boxes of the detected objects (vessels, buoys) are represented as red boxes. (c,d) yellow and gray pixels are from the X-band and W-band radar each, and the bounding boxes of detected objects are highlighted with "cyan" color.}
    \label{fig:annotation}
\end{figure*}

\subsubsection{X-Band Radar}

X-band radar data, commonly referred to as marine radar data, is stored as Cartesian coordinate image files \bl{in \texttt{PNG} format}. The radar's maximum range is set to  \bl{1 nautical miles (1,852 m) for the \texttt{Port} sequences and 1.5 nautical miles (2,778 m) for the \texttt{Island} sequences. Due to the large error bound in the X-band radar SDK, we manually refined the range estimates using satellite imagery, LiDAR, and W-band radar data. The adjusted ranges are 2,498 m for the \texttt{Port} sequence and 3,328 m for the \texttt{Island} sequence.} We fixed the sensor update rate at 1 Hz to optimize sensor fusion efficiency.
Originally, the radar required three seconds to complete a full 360-degree rotation. However, the data provided updates one-third of the full image per second. This modification was implemented to reduce dynamic changes that occur over short time intervals, thereby enhancing the temporal resolution of the sensor data.

% \begin{table}[!t]
% \caption{The radar specifications utilized in dataset}
% \label{tab:radar_config}
% \resizebox{\columnwidth}{!}{%
% \begin{tabular}{c|c|c|c|c} \toprule
% Dataset      & Manufacture & Model     & Range Resolution  & Range \\ \midrule
% Port    & SIMRAD      & HALO4 & 2.441m  & 2500m   \\
% Port    & Navtech    & RAS6-DEV-A    & 0.175m     & 600m  \\ \midrule
% Island    & SIMRAD      & HALO4 & 1.781m  & 1852m   \\
% Island    & Navtech    & RAS6-DEV-X    & 0.044m    & 330m  \\ \bottomrule
% \end{tabular}%
% }
% \end{table}

\subsubsection{W-Band Radar}

W-band radar, commonly referred to as imaging radar, was employed in our dataset using the Navtech RAS6 models. This radar system has a range of approximately 600 meters and provides data in the form of polar coordinate images. Two variants of the W-band radar were utilized for different experimental sequences. For the \texttt{Port} sequences, we used the RAS6-DEV-A model, characterized by radar rays that propagate horizontally, detecting objects at the same elevation level as the radar itself. In contrast, the \texttt{Island} sequences utilized the RAS6-DEV-X model, which emits rays capable of detecting objects below the radar's elevation. This adaptation was necessary to accommodate the vessel's size differences; specifically, the vessel in the \texttt{Island} sequences has a height of approximately 6 meters, double that of the vessel used in the \texttt{Port} sequences. \bl{For consistency, the W-band data is saved in \texttt{PNG} format for both sequences.}
Although W-band radar is traditionally utilized for ground vehicle navigation, it proves to be highly advantageous for maritime applications. It offers superior resolution in detecting surrounding environments compared to X-band radar and exhibits greater robustness than sensors based on optical wavelengths. The detailed specification of the radar configuration is detailed in \tabref{tab:radar_config}.
\begin{table}[]
\caption{Radar Specifications \bl{(S: SIMRAD, N: Navtech)}}
\label{tab:radar_config}
\resizebox{\columnwidth}{!}{%
\begin{tabular}{c|c|c|c|c|c} \toprule
 Dataset      & Type  & Model     & Resolution  & Range & Band Frequency\\ \midrule
\multirow{2}{*}{Port}  & X-band        & HALO4 (S) & 2.44 m/px  & 2498 m  &  9.410 $\sim$ 9.498 GHz \\
   & W-band     & RAS6-DEV-A (N)    & 0.175 m/px     & 600 m & 76 $\sim$ 77 GHz \\ \midrule
\multirow{2}{*}{Island}  & X-band        & HALO4 (S) & 3.25 m/px  & 3328 m  & 9.410 $\sim$ 9.498 GHz\\
  & W-band      & RAS6-DEV-X (N)   & 0.175 m/px    & 600 m & 76 $\sim$ 77 GHz \\ \bottomrule
\end{tabular}%
}
\end{table}
\subsubsection{LiDAR and Camera}
To validate the environmental conditions during sensor data acquisition, we additionally equipped the vessel with fundamental navigational sensors, LiDAR and a stereo camera. \bl{LiDAR data are stored in \texttt{BIN} files, containing the ($x$, $y$, $z$) coordinates and intensity index with \textit{float32} format.} Due to LiDAR's limited detection range, most of its data is sparse except in the berthing regions. The primary reason for incorporating LiDAR into this dataset is to assess precision during berthing maneuvers, where radar saturation can occur. 
% \bl{LiDAR data is provided in \texttt{bin} format, }

The stereo camera data, \bl{which are saved in \texttt{JPG} format}, allowed us to detect nearby vessels and structures even when operating in the open ocean. This visual data serves as ground truth for object labeling tasks. Estimating visual odometry using only the stereo camera is challenging in oceanic environments due to the scarcity of discernible objects; however, this presents an additional challenge for researchers to address.

\subsubsection{Calibration and Ground Truth}
The extrinsic calibration files and positions from the GNSS receiver are provided as text files. We employed the Hemisphere V500 GNSS receiver, which delivers precise positioning along with heading information. %While the receiver's native update frequency is 10 Hz, we have synchronized the data with other sensors to a 1 Hz update rate in our dataset.
The GNSS receiver serves as the reference frame for the vessel model, and the extrinsic parameters \bl{for other sensors are represented in this base frame.}
\bl{The GNSS-based maps are presented in \figref{fig:maps}. Both X-band and W-band radars produce accurate maps corresponding to their sensor specifications, and LiDAR generates detailed map in the narrow berthing region. However, signal reception failures occur in specific regions, leading to significant positional drift. To ensure rigorous odometry evaluation, a precisely adjusted ground truth is required. As part of future work, we aim to optimize all sensor data and generate a refined ground truth to facilitate high-precision state estimation using our dataset.}

\subsubsection{Annotation Label}
We provide the ground truth of 2D bounding box annotations for detected objects in X-band radar, W-band radar, and stereo images in the \texttt{Single Island} sequence. The annotations in the JSON file contain [$timestamp$, $id$, $category\_id$, $xmin$, $ymin$, $width$, $height$]. $id$ represents the tracked object ID that is detected in consecutive frames. The $category\_id$ is fixed at a value of 1, corresponding to the objects that should be avoided during navigation (e.g. vessels or buoys). The [$xmin$, $ymin$, $width$, $height$] parameters represent the bounding box location and dimensions of the detected objects in each image. \bl{In X-band radar data, multipath effects often result in the presence of ghost objects or terrain artifacts. To ensure reliability, we annotated only genuine objects by cross-referencing information from additional sensor modalities. The provided bounding boxes assist in identifying potential false detections within the radar data. Furthermore, occluded objects tend to merge into a single cloud. To assign a track ID to these combined vessels, we employed a weighting scheme that considers vessel size, detection frequency, and spatial consistency. Finally, due to the limited resolution of X-band radar, certain vessels may appear fused with surrounding terrain. In such cases, vessel bounding boxes were delineated separately only when their contours could be distinctly identified.} For the W-band radar, the locations of the bounding box are represented in 1024$\times$1024 Cartesian image coordinates, which are converted from the polar coordinate image. \figref{fig:annotation} depicts an example of provided 2D annotations in each images.

\subsection{Sequences}

\begin{figure*}[!t]
    \centering
    \subfigure[Global Location\label{fig:global_loc}]{
		\includegraphics[width=0.23\linewidth]{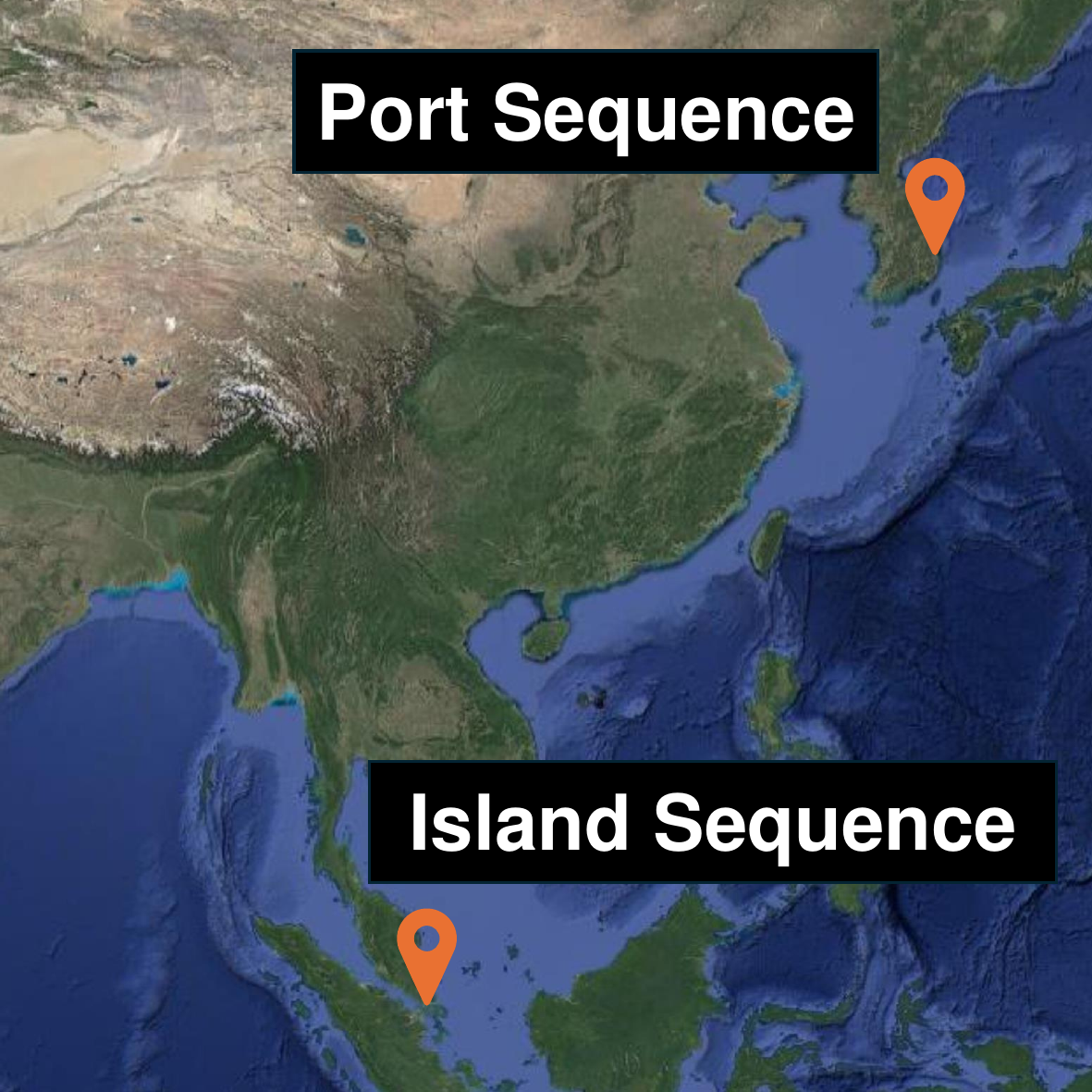}
	}
    \subfigure[Near Port\label{fig:near_port_gt}]{
		\includegraphics[width=0.23\linewidth]{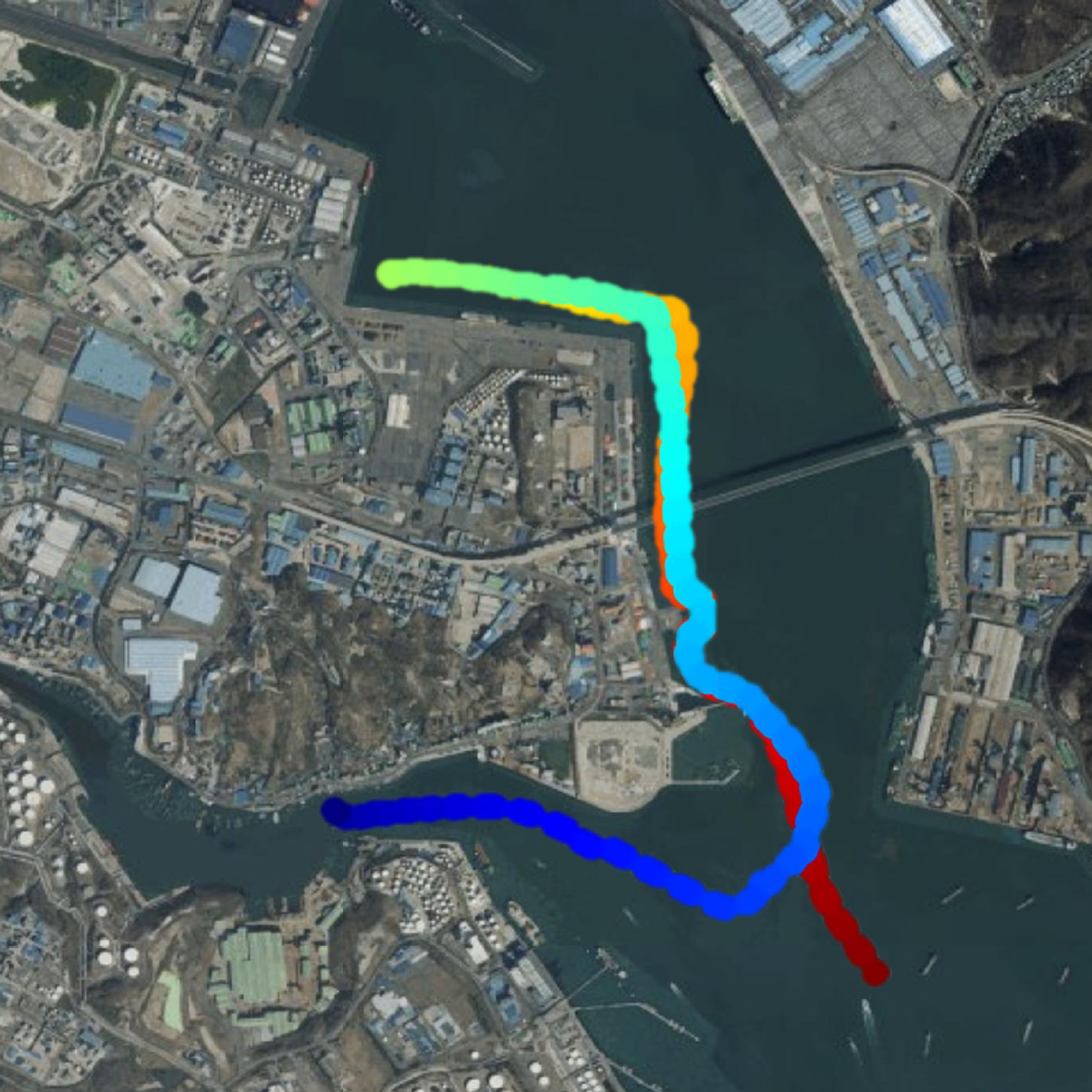}
	}
    \subfigure[Outer Port\label{fig:outer_port_gt}]{
		\includegraphics[width=0.23\linewidth]{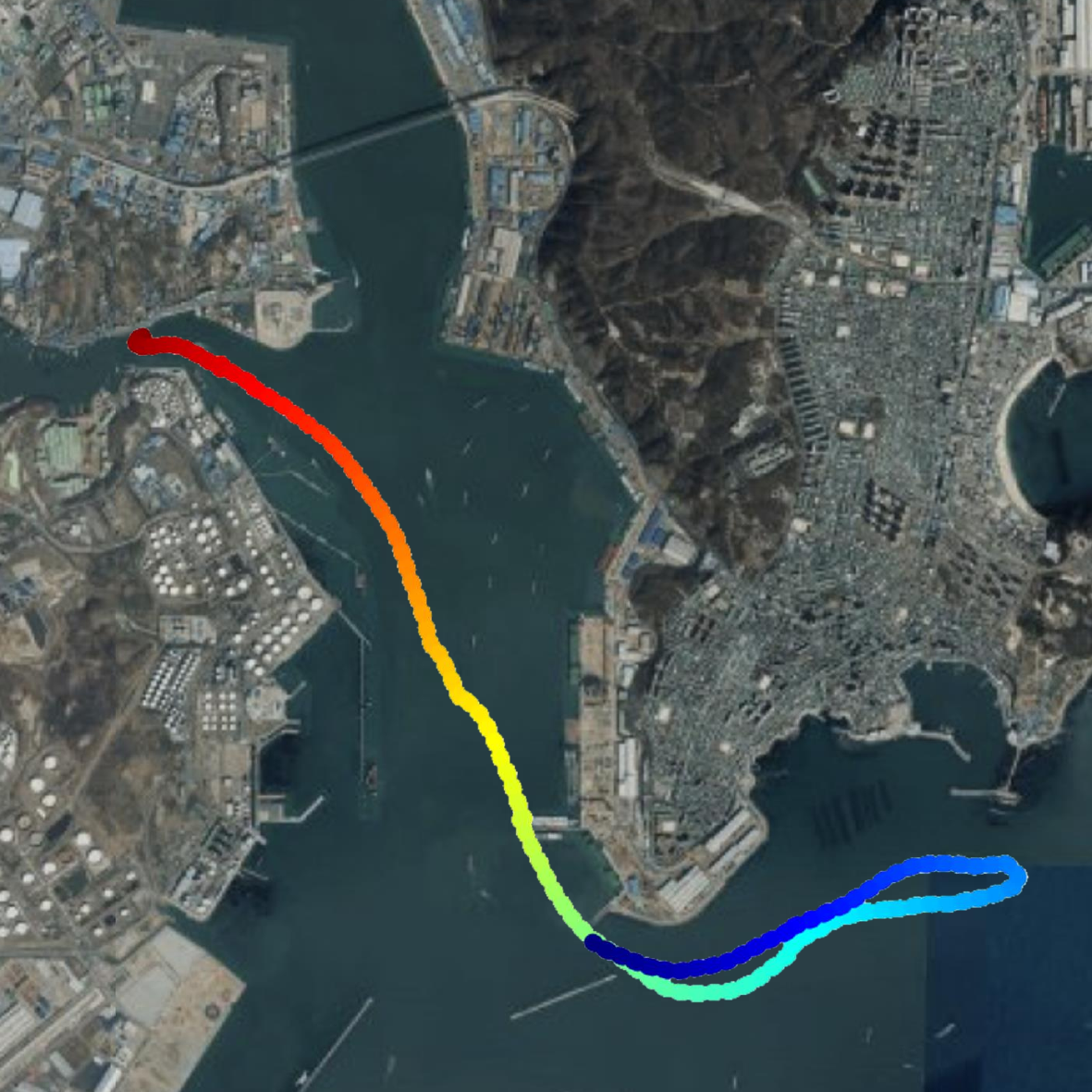}
	}
       \subfigure[Seaside\label{fig:seaside_gt}]{
		\includegraphics[width=0.23\linewidth]{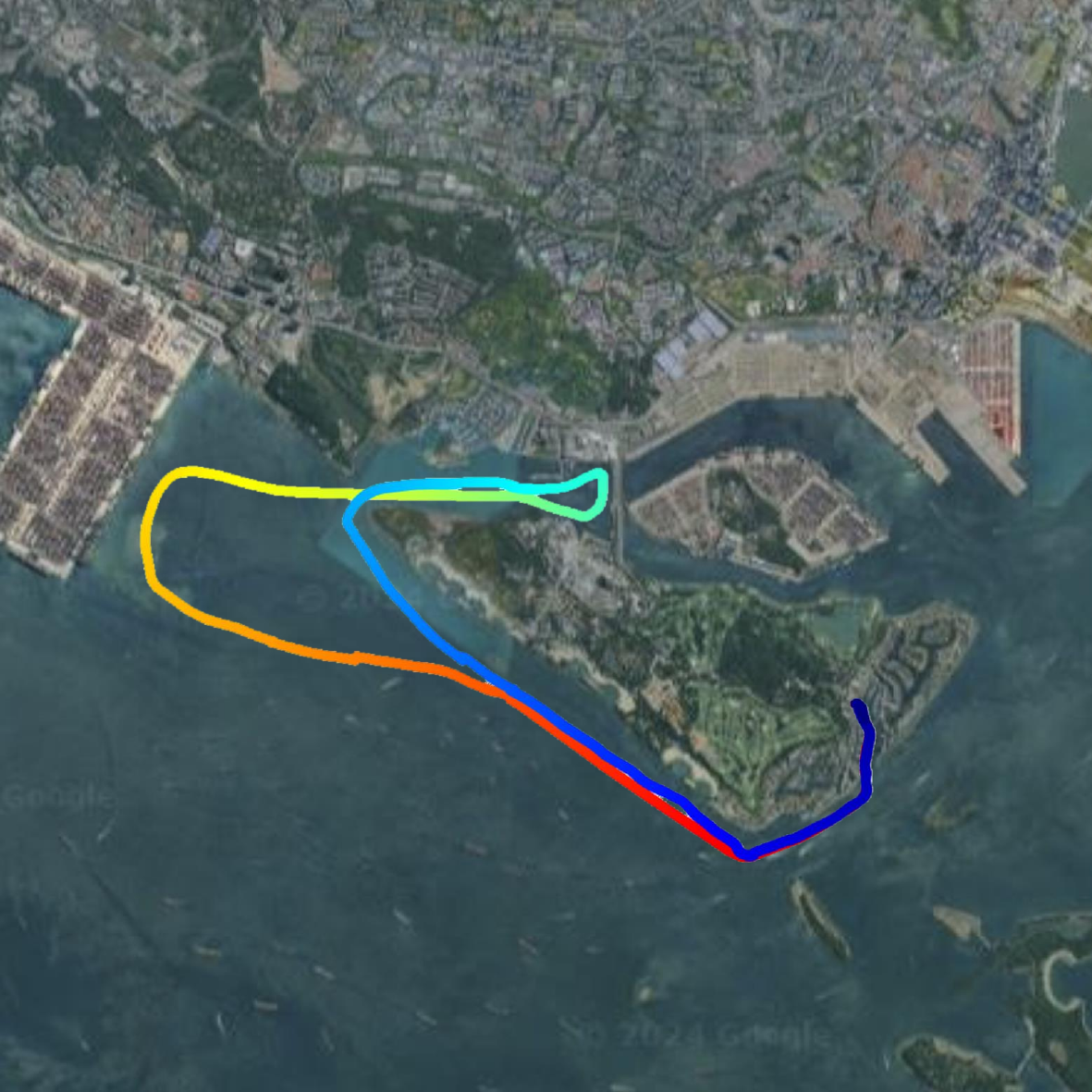}
	}
    \subfigure[Single Island\label{fig:single_island_gt}]{
		\includegraphics[width=0.23\linewidth]{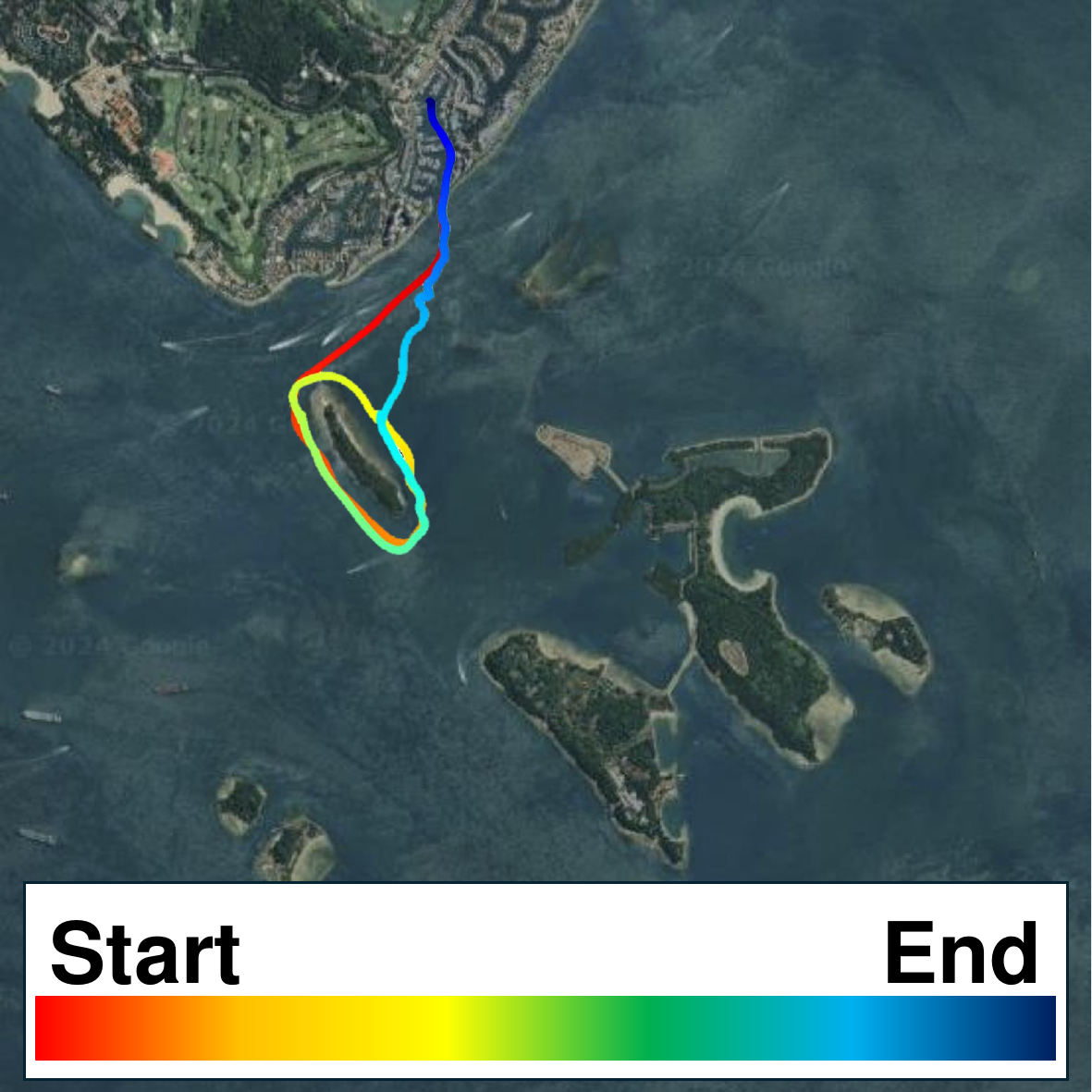}
	}
     \subfigure[Twins Island\label{fig:twin_island_gt}]{
		\includegraphics[width=0.23\linewidth]{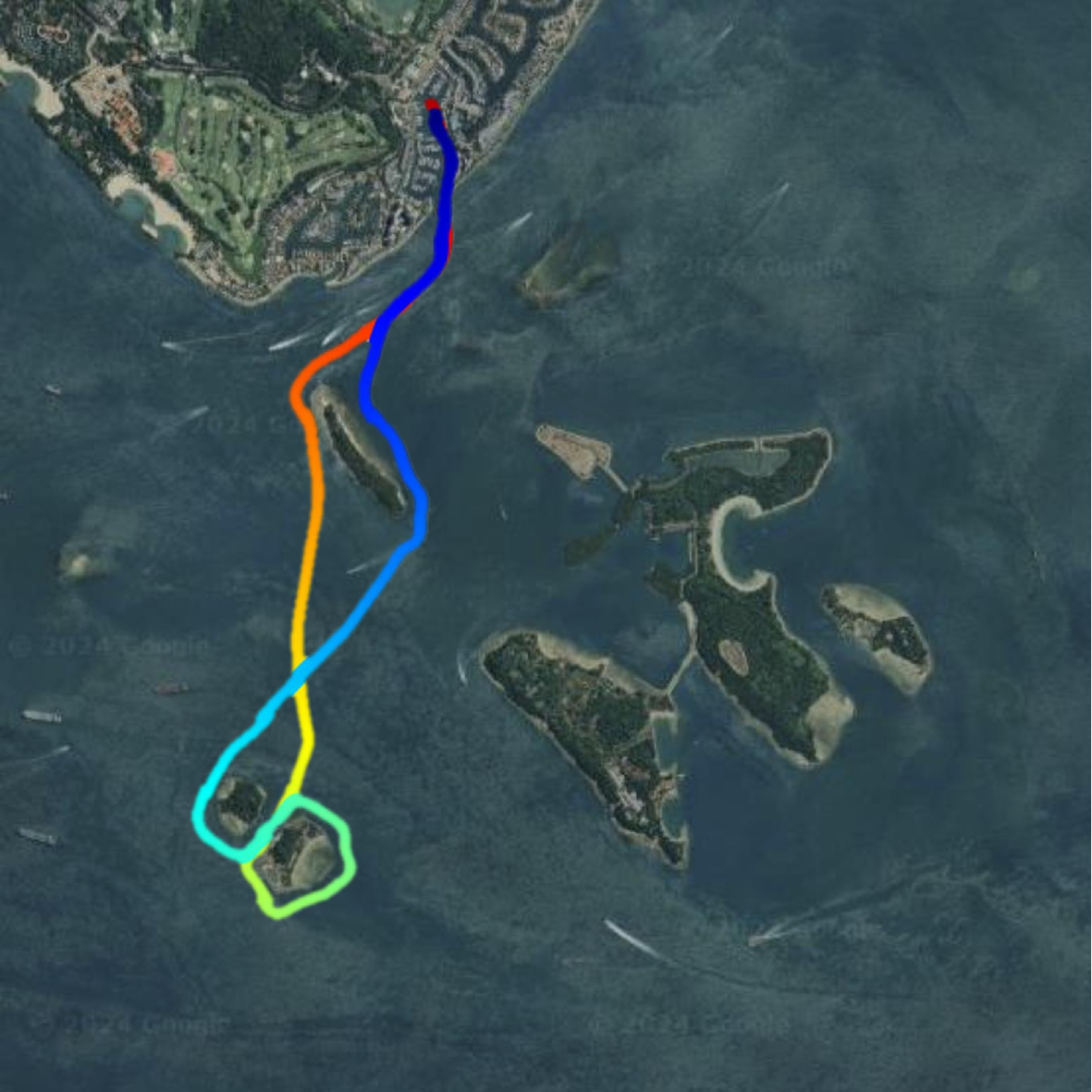}
	}
     \subfigure[Complex Island\label{fig:complex_island_gt}]{
		\includegraphics[width=0.23\linewidth]{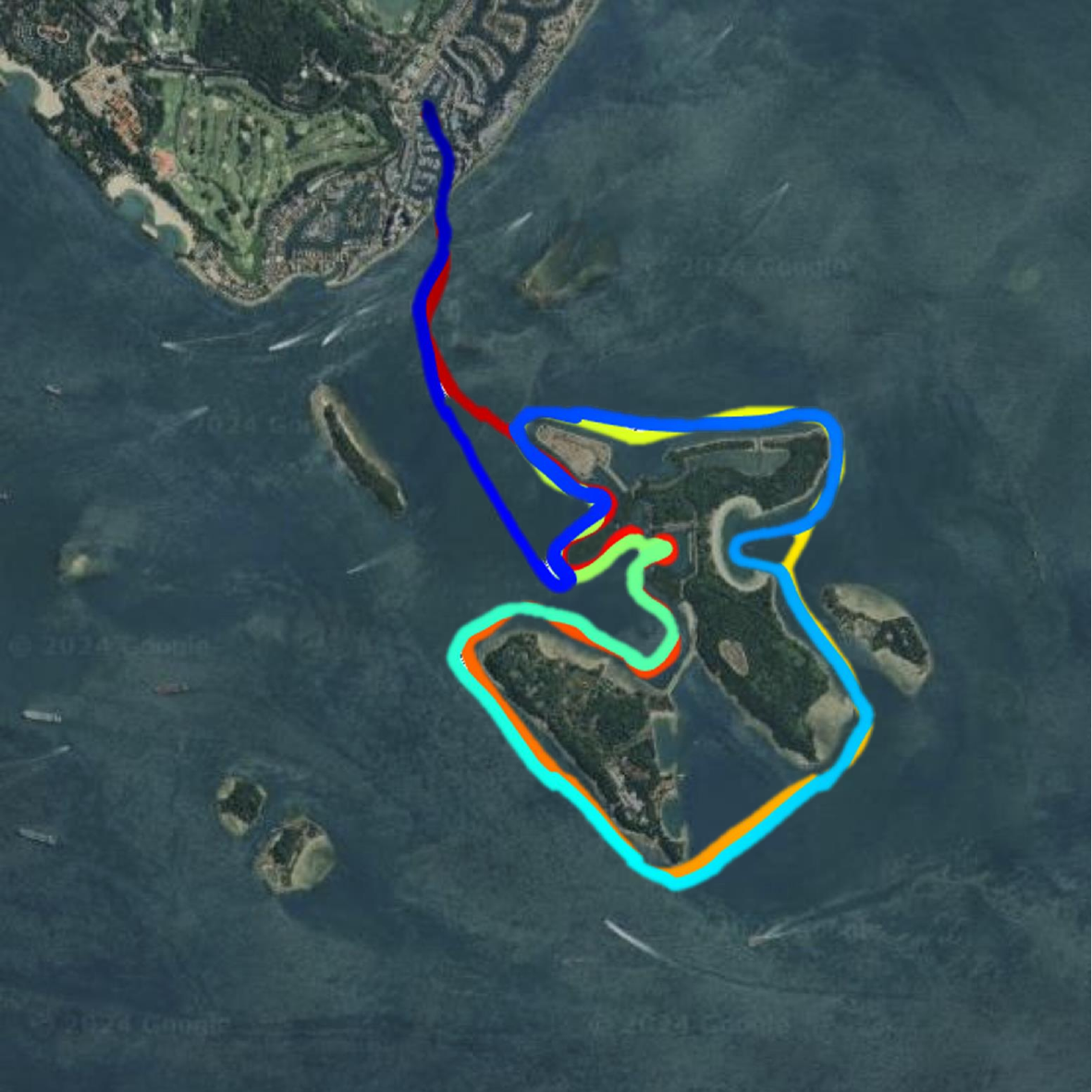}
	}
     \subfigure[Island Expedition\label{fig:island_expedition_gt}]{
		\includegraphics[width=0.23\linewidth]{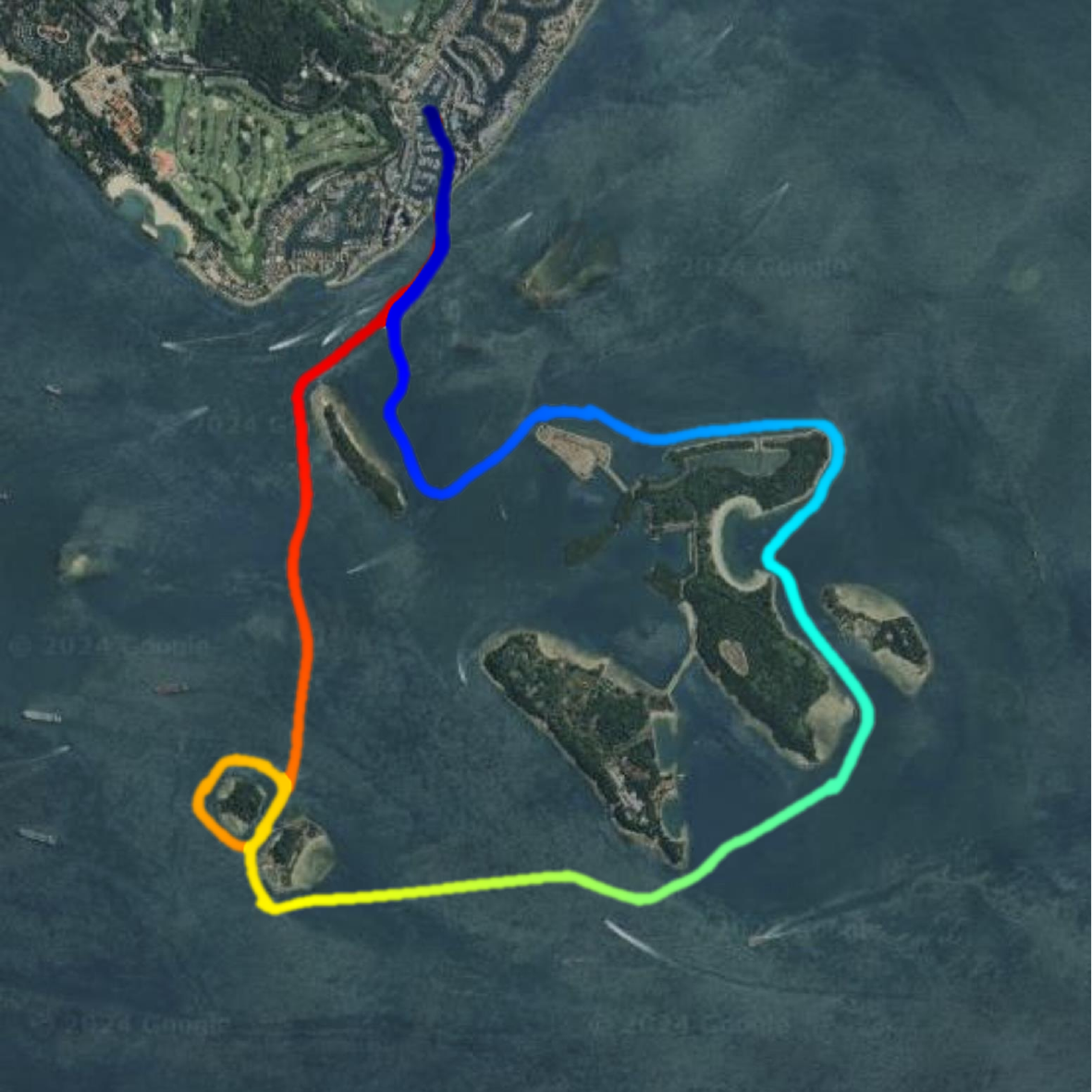}
	}
    \caption{Ground truth trajectories for all sequences. The \texttt{Port} sequences (b, c) were collected in South Korea, while the \texttt{Seaside} and \texttt{Island} sequences (d, e, f, g, h) were acquired in Singapore. All sequences share \bl{slight} common locations except for the \texttt{Island Expedition} sequence (h), which encompasses three sub-sequences (e, f, g). Trajectories commence in red and terminate in blue.}
    \label{fig:gt}
\end{figure*}

The MOANA dataset encompasses two distinct regional environments: South Korea and Singapore. The data collected in South Korea is from Ulsan, an area characterized by predominantly industrial structures. The port area is developed for cargo transport, and the surroundings are densely packed with factory sites and artificial structures. This provided a highly structured environment, which we refer to as the \texttt{Port} sequence.

In contrast, the dataset from Singapore is collected in the Harborfront area, representing a largely unstructured environment. Apart from the berthing point, the surroundings are dominated by natural elements such as trees and rocks, making localization algorithms more challenging. Given the varying sizes of islands in this region, we have designated this sequence as the \texttt{Island} sequence. The overall trajectory for each sequence is depicted in \figref{fig:gt}, with detailed explanations of each sequence and the associated challenges provided below.

\subsubsection{Port Sequence: }
The \texttt{Port} sequences represent a highly structured, industrial environment characterized by strong and consistent detections of the surrounding area. This sequence is primarily designed for odometry or \ac{SLAM} applications, where generating a reliable map of the port is the main objective.
The GNSS data exhibits periodic delays due to the sensor interference and regional characteristics in the \texttt{Port} sequences, highlighting the critical importance of achieving accurate sensor-driven \ac{SLAM} results.

These sequences are the most straightforward for algorithm testing, with the primary challenge being the mitigation of wave-induced wobble, which leads to inconsistencies in radar data. As radar systems rely on reflected signals, this wobble can disrupt the \ac{RCS} continuity of sensor measurements.
Additionally, the presence of numerous anchored large vessels introduces both opportunities and challenges for tracking. The radar-reflective coatings on these ships make them ideal objects for tracking in open water. However, proximity to these vessels can lead to significant multipath effects, complicating the tracking process.

% In the \texttt{Port} sequences, sensor interference and regional characteristics resulted in the GNSS data exhibiting periodic delays.

\textbf{(\textit{i}) \texttt{Near Port}: }
In the \texttt{Near Port} sequence, the dataset primarily emphasizes W-band radar due to its superior performance in short-range detection, as X-band radar exhibits limitations in this condition. W-band radar is, therefore, the predominant sensor for capturing the surrounding environment in these sequences. A significant challenge arises \bl{at the midpoint of the entire route}, where severe multipath effects generate ghost objects, a critical obstacle to achieving robust navigation.

\textbf{(\textit{ii}) \texttt{Outer Port}: }
In contrast, the \texttt{Outer Port} sequence focuses on X-band radar, as W-band radar's limited detection range renders it ineffective in certain areas. For both odometry and mapping tasks, X-band radar serves as the primary sensor. However, in the narrow-loop regions, W-band radar can be effectively utilized, provided its characteristics are leveraged appropriately.

\subsubsection{Island Sequence: }

\bl{The \texttt{Island} sequences depict an unstructured environment characterized by sparse, irregular radar returns predominantly arising from natural elements.} Due to these features, surrounding detections are inconsistent, making it challenging to recognize the same locations reliably.
The dataset includes three distinct island environments: single island, twins island, and complex island. Each sequence contains overlapping areas with others, facilitating global place recognition across the dataset.
In addition to island areas, the sequence also includes seaside areas, providing a mix of both unstructured natural environments and more structured coastal regions.

\textbf{(\textit{i}) \texttt{Single Island}: }
The \texttt{Single Island} sequence features the shortest route in the island dataset, consisting of a simple loop around the island collected at night. The yacht completes two laps before returning to the berthing point.

\textbf{(\textit{ii}) \texttt{Twins Island}: }
\texttt{Twins island} is located at the farthest distance and experiences sporadic data loss in the W-band radar. However, as the vessel navigates through the narrow waterway between the islands, W-band radar becomes essential for robust positioning and detection, compensating for the limitation of the X-band radar. \texttt{Twins island} sequence begins in darkness and gradually transitions to brightness by the end.

\textbf{(\textit{iii}) \texttt{Complex Island}: }
\texttt{Complex island} is the largest island, which is suited for evaluating odometry algorithms. Long, circular route makes the distance between the starting and end points a valuable metric for assessing marine odometry performance.

\textbf{(\textit{iii}) \texttt{Island Expedition}: }
\texttt{Island Expedition} encompasses all three islands but only includes partial segments from each. This dataset is ideal for testing global localization algorithms.

\textbf{(\textit{iv}) \texttt{Seaside}: }
Lastly, the \texttt{Seaside} sequence incorporates both beachside environments and a container loading zone. This sequence is particularly useful for analyzing the impact of both structured and unstructured environments within a single dataset. \texttt{Seaside} sequence starts in bright daylight and ends in darkness.

\begin{table*}[t!]
\caption{Absolute Trajectory Error for W-band and X-band radars. Due to challenging data in other sequences, only the \texttt{Near Port} sequence was evaluated using both W-band and X-band radar odometry algorithms. The W-band radar demonstrates lower error in the \texttt{Near Port} sequence, attributable to its superior short-range detection accuracy compared to the X-band radar. In contrast, the X-band radar exhibits robust performance across all other sequences not covered by the W-band radar sensor.}
\resizebox{\linewidth}{!}{
\begin{tabular}{c|c|cc|ccccc} \toprule
\multirow{2}{*}{Method} &
\multirow{2}{*}{Radar} &
  \multicolumn{2}{c|}{Port Sequence} &
  \multicolumn{5}{c}{Island Sequence} \\ 
 &
 &
  Near Port &
  Outer Port &
  Single Island &
  Twins Island &
  Complex Island &
  Island Expedition &
  Seaside
   \\ \midrule
\begin{tabular}[c]{@{}c@{}}\textbf{CFEAR}\\ \cite{adolfsson2022lidar}  \end{tabular} & W-Band  & 20.35  & - & - & - & - & -  & - \\  \hline
\begin{tabular}[c]{@{}c@{}}\textbf{LodeStar}\\ \cite{jang2024lodestar} \end{tabular} & X-Band & 63.48 & 78.92 & 41.59 & 53.17 & 22.88 & 130.48 & 35.81 \\  \hline
\end{tabular}}

\label{tab:benchmark}
\end{table*}

\begin{figure*}[!t]
    \centering
    \subfigure[Near Port (W-band)]{
		\includegraphics[width=0.45\columnwidth]{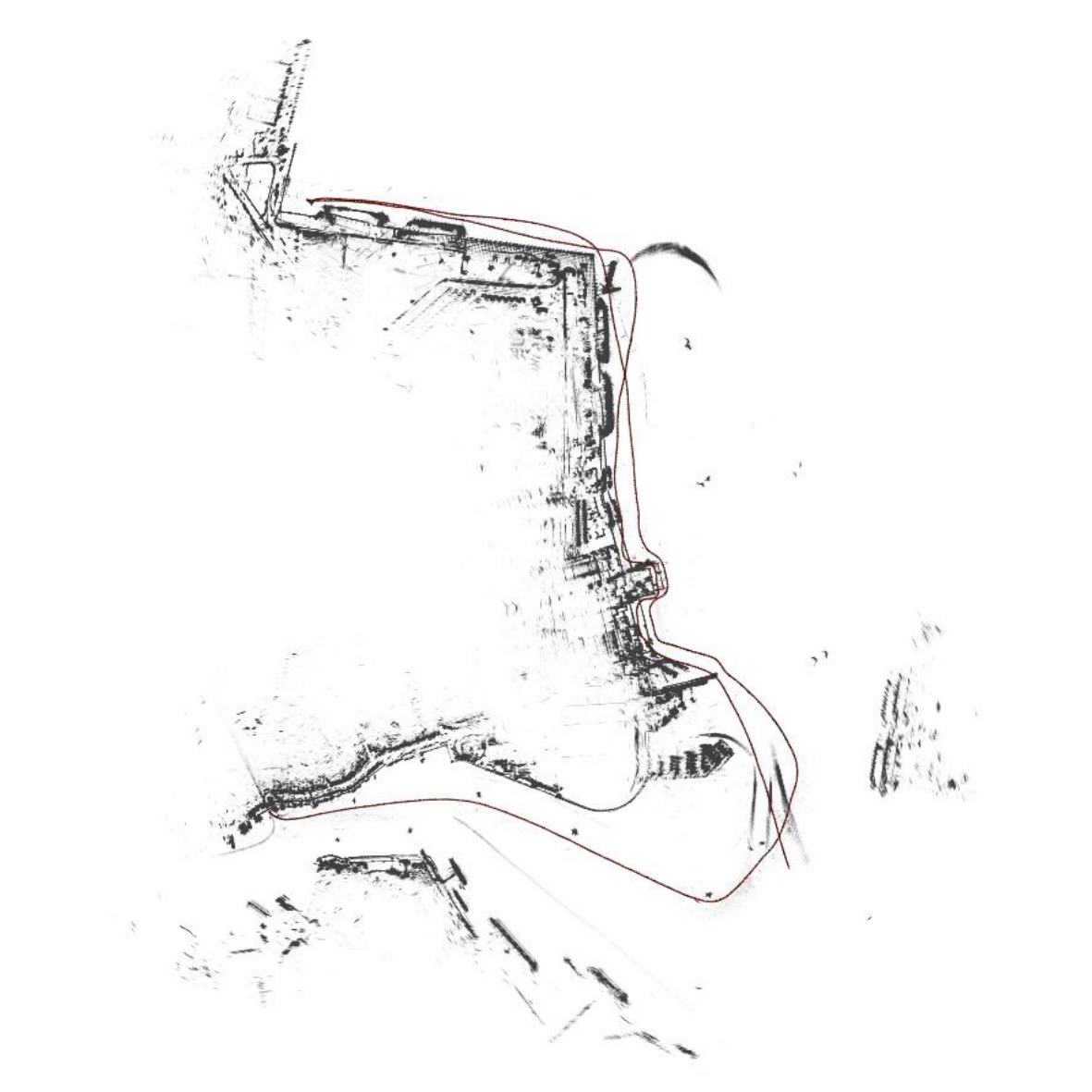}
	}
    \subfigure[Near Port (X-band)]{
		\includegraphics[width=0.45\columnwidth]{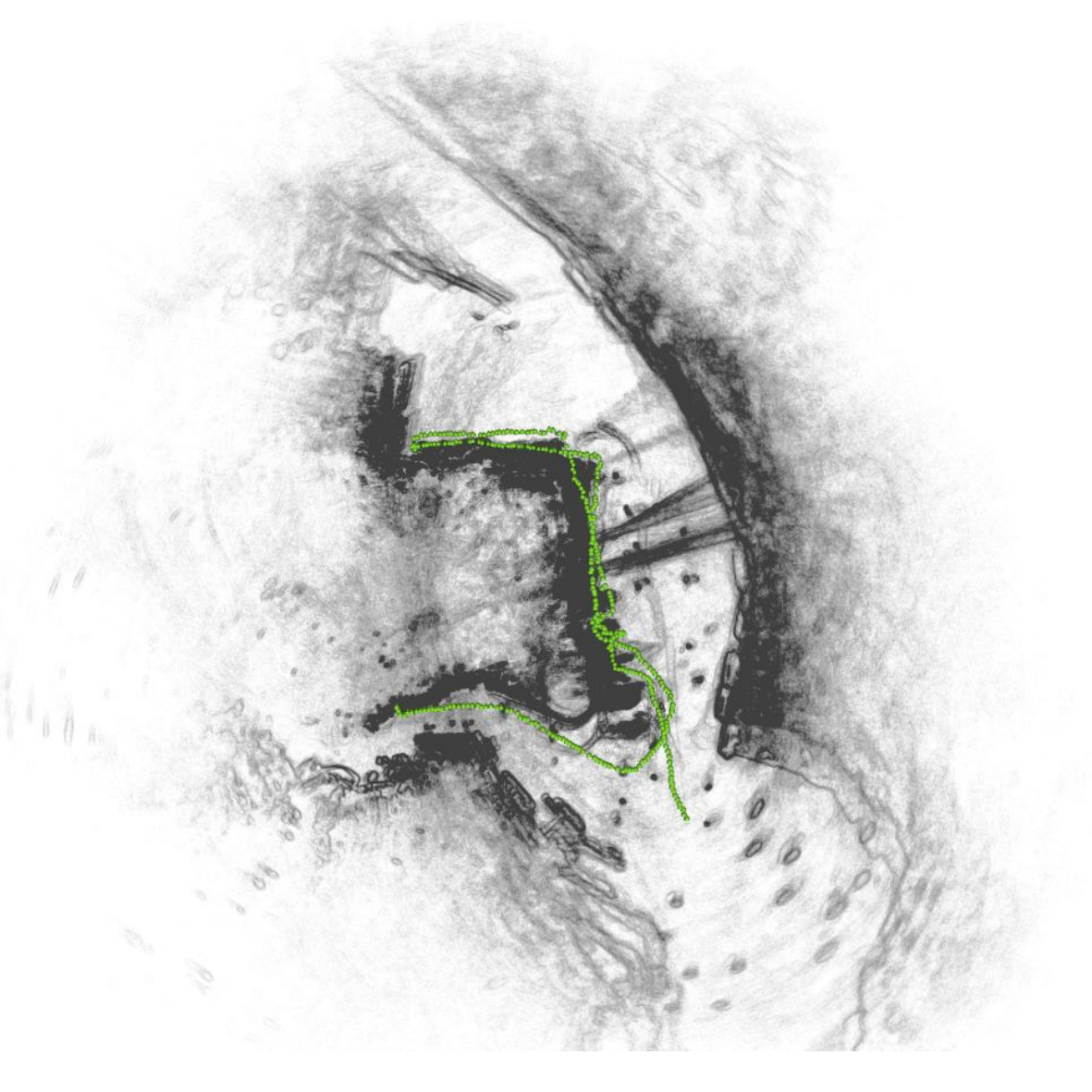}
	}
 \subfigure[Outer Port (X-band)]{
		\includegraphics[width=0.45\columnwidth]{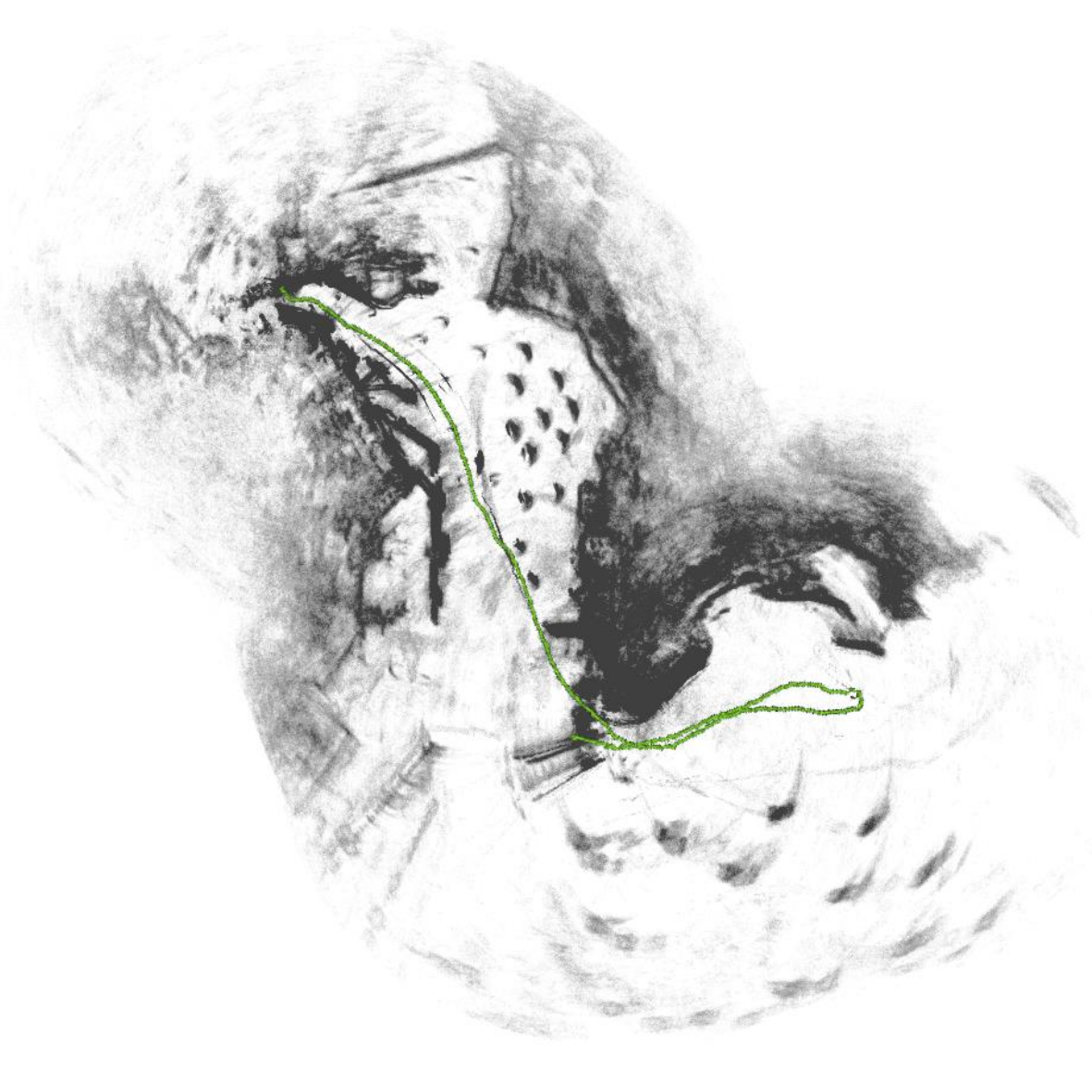}
	}
 \subfigure[Seaside (X-band)]{
		\includegraphics[width=0.45\columnwidth]{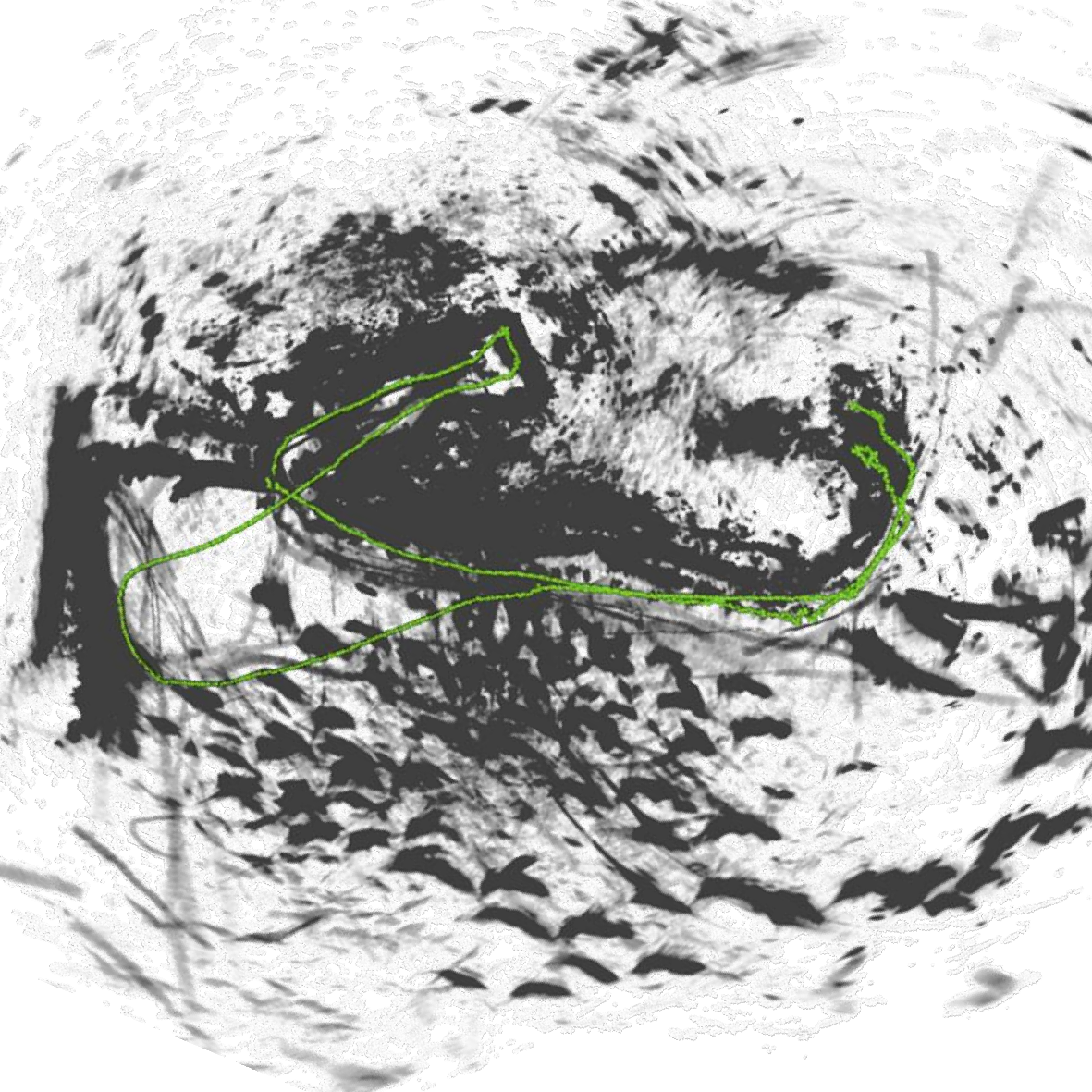}
	}\\
 \subfigure[Single Island (X-band)]{
		\includegraphics[width=0.45\columnwidth]{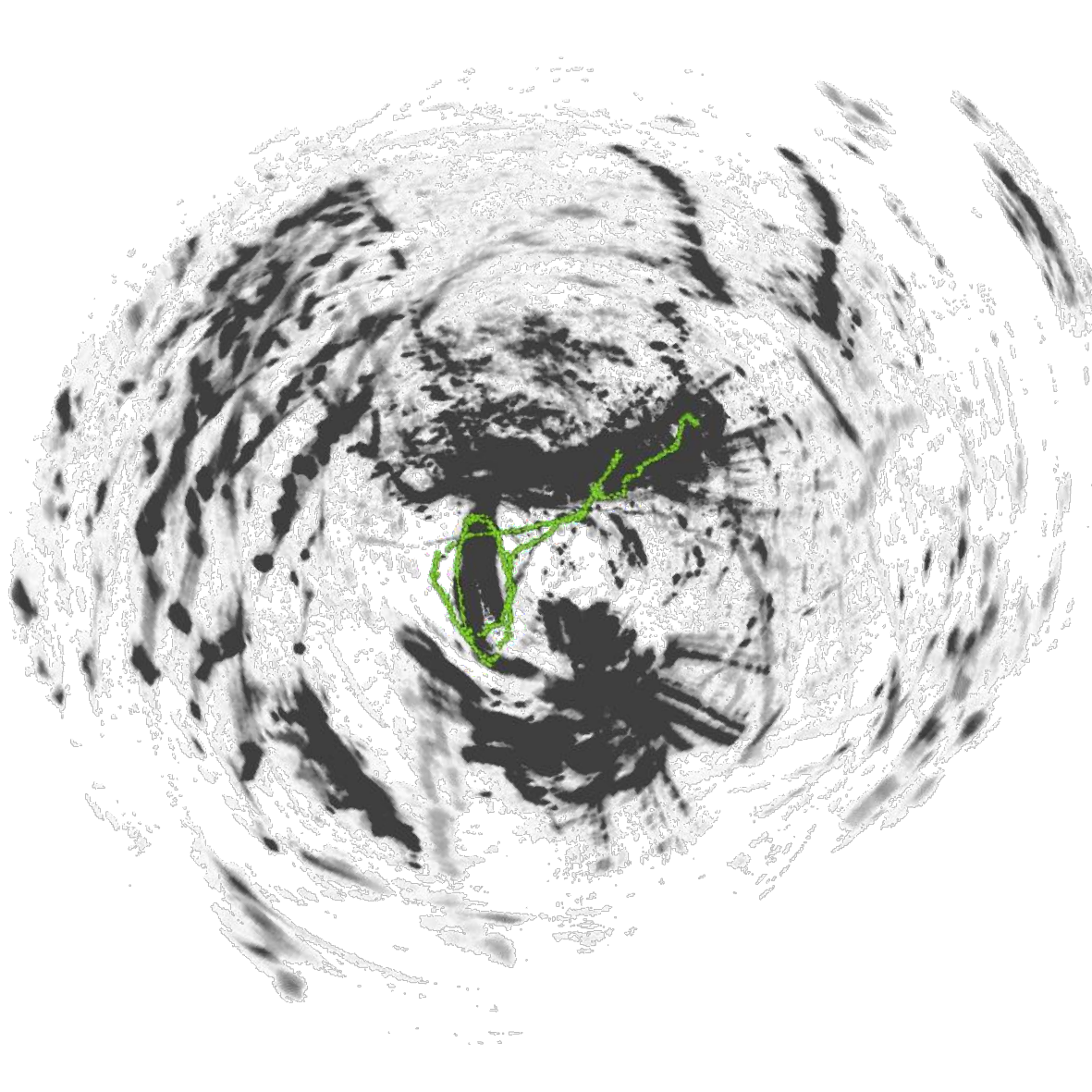}
	}
 \subfigure[Twins Island (X-band)]{
		\includegraphics[width=0.45\columnwidth]{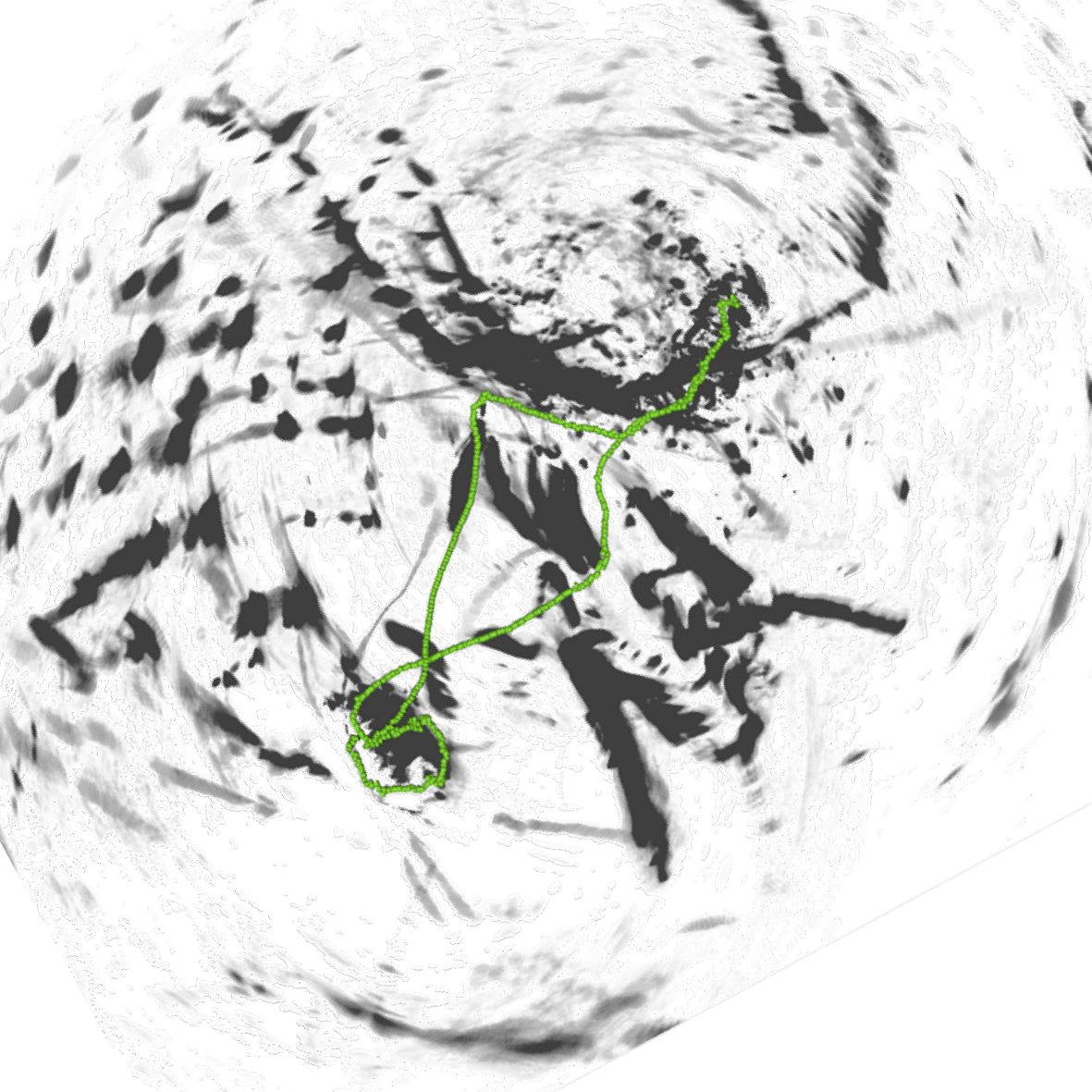}
	}
 \subfigure[Complex Island (X-band)]{
		\includegraphics[width=0.45\columnwidth]{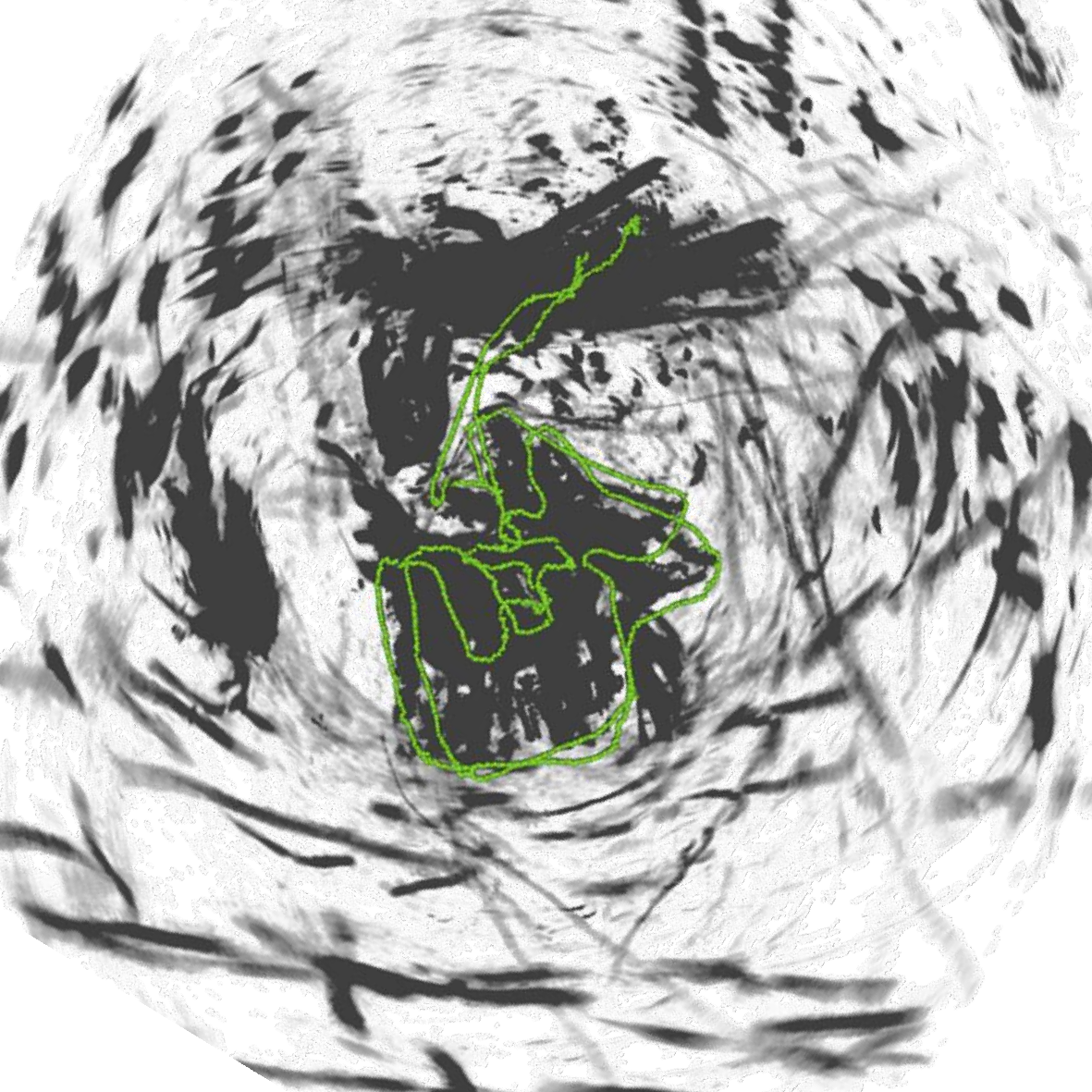}
	}
    \subfigure[Island Expedition (X-band)]{
		\includegraphics[width=0.45\columnwidth]{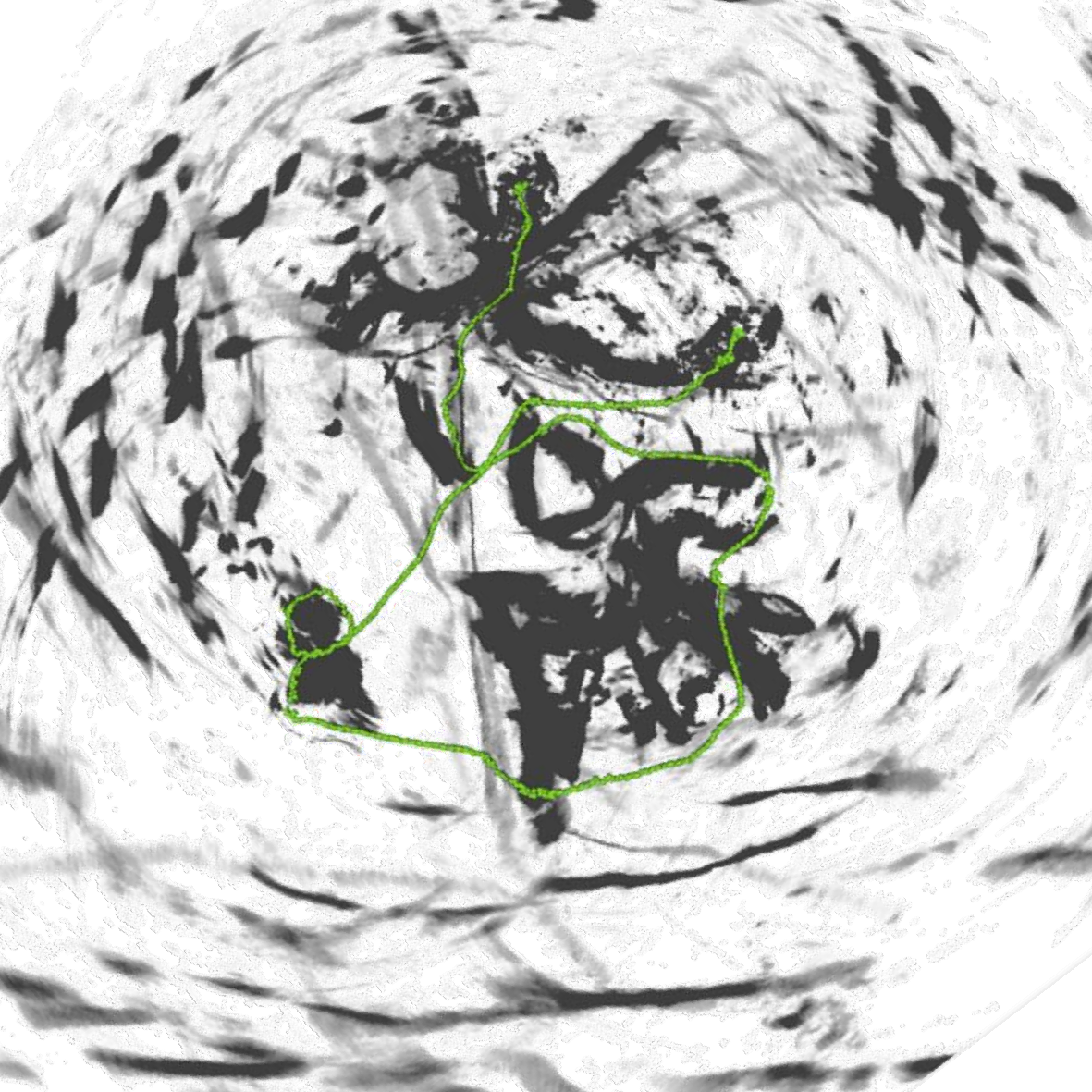}
	}
    \caption{W-band and X-band radar odometry/mapping results. The \texttt{Port} sequences achieve promising odometry performance due to the presence of continuous features such as walls and bridges. In contrast, the \texttt{Island} sequences, which lack these features, show insufficient performance in W-band radar odometry. However, X-band radar still demonstrates superior performance compared to W-band radar odometry. In near-land areas, adjusting the vessel’s pose using W-band radar could further enhance odometry accuracy.}
    \label{fig:slam}
\end{figure*}

\section{Radar Odometry Benchmark}

\subsection{W-band radar odometry (CFEAR)}
For the benchmark test of W-band radar odometry, we employed the CFEAR~\citep{adolfsson2022lidar} method, a state-of-the-art approach for W-band radar odometry estimation. Although our goal was to evaluate the entire trajectory, \bl{lacking distinct objects in certain regions} made this infeasible. As a result, the W-band radar odometry had to be limited to sections where the vessel remained near the shoreline. The results are depicted in \figref{fig:slam} and \tabref{tab:benchmark}. 

\begin{figure*}[!t]
    \centering
 \includegraphics[width=0.8\linewidth]{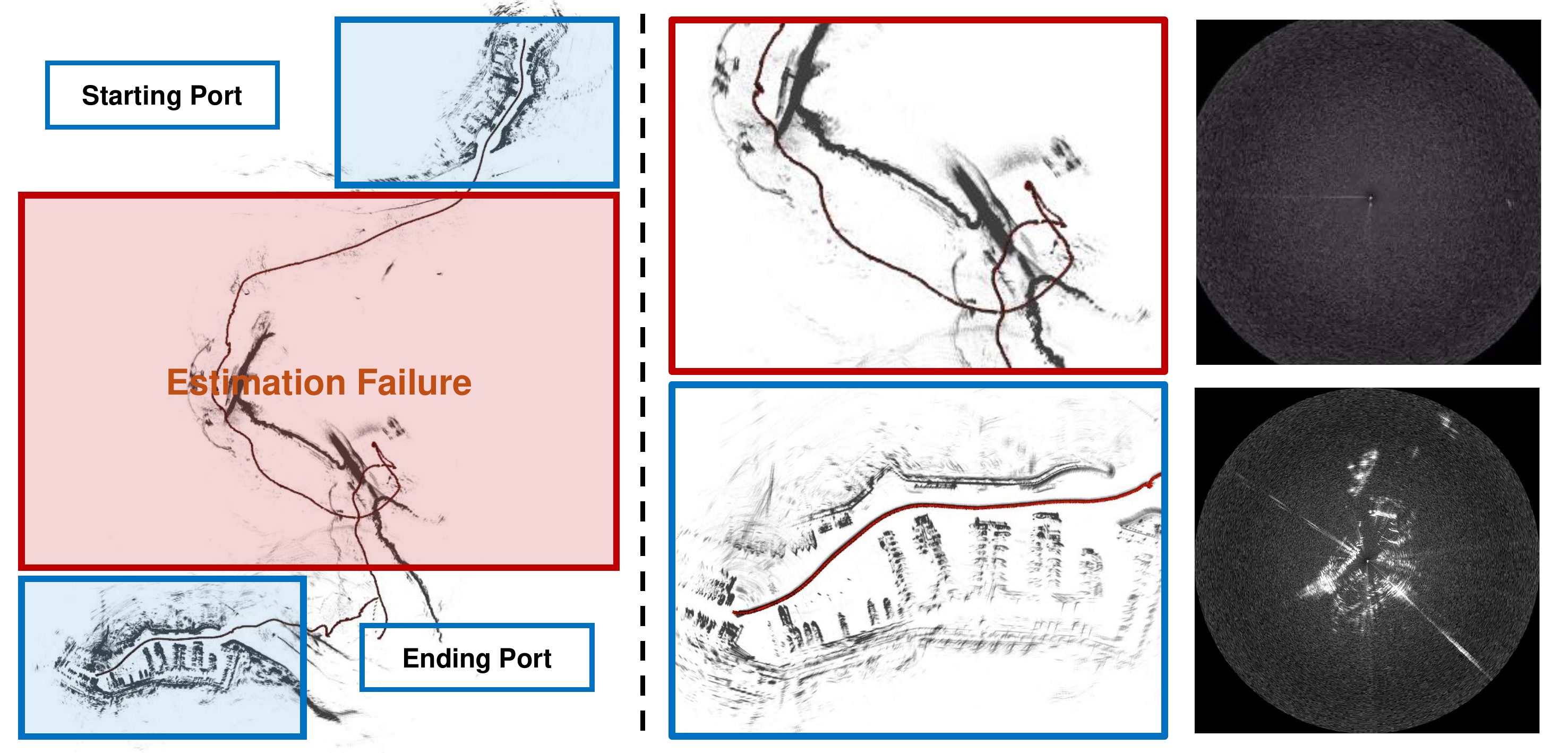}
    \caption{Odometry estimation results using W-band radar for the \texttt{Single Island} sequence. The W-band radar successfully generates reliable odometry and mapping outcomes with existing methods near the berthing area (blue) of the \texttt{Single Island} sequence. However, tracking loss frequently occurs when processing unstructured and featureless data (red).}
    \label{fig:cfear_results}
\end{figure*}

In the \texttt{Port} sequence, we achieved reasonable odometry estimation from \texttt{Near Port} data. However, as intended, the majority of W-band radar images in the \texttt{Outer Port} data were blank, precluding meaningful results from this sequence. This design choice highlights the necessity of incorporating additional sensors to overcome inherent limitations and ensure stable vessel navigation.
For the \texttt{Island} sequence, there were sporadic challenges that makes difficult to conduct odometry estimation. we provide test figure in \figref{fig:cfear_results} to depict our traversal around the \texttt{Single Island} with W-band radar odometry estimation. While berthing area features were successfully captured, ambiguous detection of unstructured environments such as trees and sand hindered frame-to-frame matching, presenting challenges that warrant further investigation.
In summary, we observed that the W-band radar can accurately depict surrounding environments, but its short-range limitation impedes the observation of movement between islands. This result underscores the need for sensor fusion with longer-range sensors, which are included in this dataset. Integrating data from the X-band radar could facilitate continuous vessel motion tracking.
% fusion 필요

\subsection{X-band radar odometry (LodeStar)}
For X-band radar odometry estimation, we employed the LodeStar~\citep{jang2024lodestar} algorithm, which is a state-of-the-art method in this domain. Unlike the W-band radar benchmark test, we utilized the full dataset for the X-band radar benchmark since the surrounding data were comprehensively captured. The only significant challenges arose from data saturation when the vessel approached too closely to territorial regions, causing multipath noise and false alarms in the sensor images.

In the \texttt{Port} sequence data, we observed robust odometry estimation despite some radar saturation occurring in the \texttt{Near Port} sequence. However, the saturation issues in the \texttt{Island} sequence were not managed, leading to tracking loss during odometry estimation. To achieve more accurate and robust odometry, incorporating near-range data from other sensors such as W-band radar can provide enhanced results.

%\subsection{Radar SLAM}
%\gr{\lipsum[10]}\gr{\lipsum[10]}\gr{\lipsum[10]}

% \begin{table*}[!t]
% \resizebox{\linewidth}{!}{
% \begin{tabular}{c|c|c|l} \toprule
% Sensor          & Manufacture & Model         & Specification \\ \midrule
% X-band Radar    & SIMRAD      & HALO4         & Range 1852m / Res. 3m  \\
% W-band Radar    & Navtech     & RAS6          & Range 600m / Res. 0.175m \\
% LiDAR           & Robosense   & RS-Helios-5515 & Range 150m / HFOV 360$^{\circ}$, VFOV 70$^{\circ}$ \\
% Stereo Camera   & Seadronix    & -             & 12 Mega Pixel / Res. 2028x1520 / HFOV 160$^{\circ}$, VFOV 160$^{\circ}$  \\
% RTK-GNSS        & NavtechGPS  & Hemisphere V500 & Acc. 8mm \\ \bottomrule
% \end{tabular}%
% }
% \end{table*}

\section{Radar Object Detection in Maritime Environments}

Unfortunately, there is a lack of publicly available object detection methods for both X-band and W-band radar in the maritime environment. Typically, existing object detection algorithms in the W-band radar are primarily designed for ground vehicles or pedestrian detection in land environments, which are not adaptable for marine vessel detection tasks. As the MOANA dataset demonstrates the potential application of W-band radar for short-range detection in maritime navigation tasks, we expect that our dataset will contribute to developing oceanic radar object detection algorithms.

\section{Conclusion}
MOANA presents the first maritime multi-radar dataset incorporating scanning radars of different bandwidths.
Our dataset facilitates the use of W-band radar in oceanic environments while maintaining vessel ego-motion tracking by leveraging existing X-band radar.
% Addressing the limitations of W-band radar, W-band radar also mitigates the vulnerability of X-band radar during berthing operations.
% Our benchmark result also demonstrated that integrating the X-band and W-band radar can further enhance the performance of maritime navigation tasks such as berthing, sailing, and docking.
% \bl{Fusion is good. Short-rage = navtech + longer range = marine. Our benchmark result yields that}
\bl{The high-resolution imaging provided by the W-band radar appears to mitigate the perceptual limitations of the X-band radar during operations requiring precise sensing capabilities, such as berthing or docking.}
\bl{Our benchmark results indicate that while W-band radar delivers enhanced odometry estimation during near harbor sequences, its performance in wide-area scenarios is limited. Thus, a hybrid approach integrating X-band and W-band radar data could potentially improve maritime navigation tasks in a complementary manner.}
Seven sequences in this dataset are expected to support advancements in autonomous navigation systems.
In future updates, we aim to extend the dataset with sequences incorporating temporal variance to evaluate robustness under time-differentiated conditions.

\begin{acks}
\bl{This work was supported by the National Research Foundation of Korea (NRF) grant funded by the Korea government (MSIT) (No. RS-2024-00461409)}, and the research in the Singapore region was supported by Defence Science and Technology Agency (DSTA)
\end{acks}

% \balance
% \small
\bibliographystyle{SageH}
\bibliography{reference}

\end{document}